\documentclass[]{cit_lab_mfr}

\usepackage{hyperref}
\usepackage{cleveref}
\usepackage{verbatim}

\usepackage{graphicx}
\usepackage{wrapfig}   % 文字环绕图/表
\usepackage{floatrow}
\usepackage{subcaption}
\usepackage{listings}
\usepackage{algorithm}
\usepackage{subcaption} % 
\usepackage[toc,page,header]{appendix}


\definecolor{diy_pink}{RGB}{255,247,240}
\usepackage{multirow}          % 你已经提示需要它
\usepackage[table]{xcolor}     % 提供 \rowcolor 等表格着色命令

\usepackage{graphicx}   % for \resizebox
\usepackage{adjustbox}  % for adjustbox 环境（推荐）
\usepackage{array}      % 自定义列格式时有用
\usepackage{graphicx}
\usepackage{caption} % or \usepackage{capt-of}
\usepackage{xcolor}
\usepackage{pifont}

\usepackage{minitoc}

\title{\textbf{\textit{FysicsWorld}}: A Unified Full-Modality Benchmark for Any-to-Any Understanding, Generation, and Reasoning}

\author{%
\parbox{\textwidth}{\centering
Yue Jiang$^{*}$,\,
Dingkang Yang$^{*,\dagger, \S}$,\,
Minghao Han$^{*}$,\,
Jinghang Han,\,
Zizhi Chen, \\
Yizhou Liu,\,
Mingcheng Li,\,
Peng Zhai,\,
Lihua Zhang$^{\S}$
}}

\affiliation{%
\parbox{\textwidth}{\centering\small
College of Intelligent Robotics and Advanced Manufacturing, Fudan University\\[1mm]
Fysics Intelligence Technologies Co., Ltd. (Fysics AI)\\[1mm]
}}

\contribution[*]{Equal contribution}
\contribution[\dagger]{Project lead}
\contribution[\S]{Corresponding Author}

\renewcommand{\thefootnote}{\fnsymbol{footnote}}
\renewcommand{\thefootnote}{\arabic{footnote}}  % Reset for body

\abstract{
Despite rapid progress in multimodal large language models (MLLMs) and emerging omni-modal architectures, current benchmarks remain limited in scope and integration, suffering from incomplete modality coverage, restricted interaction to text-centric outputs, and weak interdependence and complementarity among modalities. To bridge these gaps, we introduce \textbf{\textit{FysicsWorld}}, the first unified full-modality benchmark that supports bidirectional input–output across image, video, audio, and text, enabling comprehensive any-to-any evaluation across understanding, generation, and reasoning. 
\textbf{\textit{FysicsWorld}} encompasses 16 primary tasks and 3,268 curated samples, aggregated from over 40 high-quality sources and covering a rich set of open-domain categories with diverse question types.
We also propose the Cross-Modal Complementarity Screening (CMCS) strategy integrated in a systematic data construction framework that produces omni-modal data for spoken interaction and fusion-dependent cross-modal reasoning.
Through a comprehensive evaluation of over 30 state-of-the-art baselines, spanning MLLMs, modality-specific models, unified understanding–generation models, and omni-modal language models, \textbf{\textit{FysicsWorld}} exposes the performance disparities and limitations across models in understanding, generation, and reasoning. Our benchmark establishes a unified foundation and strong baselines for evaluating and advancing next-generation full-modality architectures.
}

\date{\today}
\checkdata[Corresponding]{\url{dicken@fyscis.ai,  jiangyue23@m.fudan.edu.cn, lihuazhang@fudan.edu.cn}}
\checkdata[Project Page]{\url{https://github.com/Fysics-AI/FysicsWorld}}
\checkdata[Dataset]{\url{https://huggingface.co/datasets/Fysics-AI/FysicsWorld}}

\begin{document}
\maketitle

\begingroup
\renewcommand{\thefootnote}{\fnsymbol{footnote}}
% \footnotetext[3]{Work was done during their internship.}
\endgroup

%%%%%% Introduction %%%%%%
\section{Introduction}
\label{sec:intro}

\begin{figure}[htbp]
    \centering
    \includegraphics[width=\textwidth]{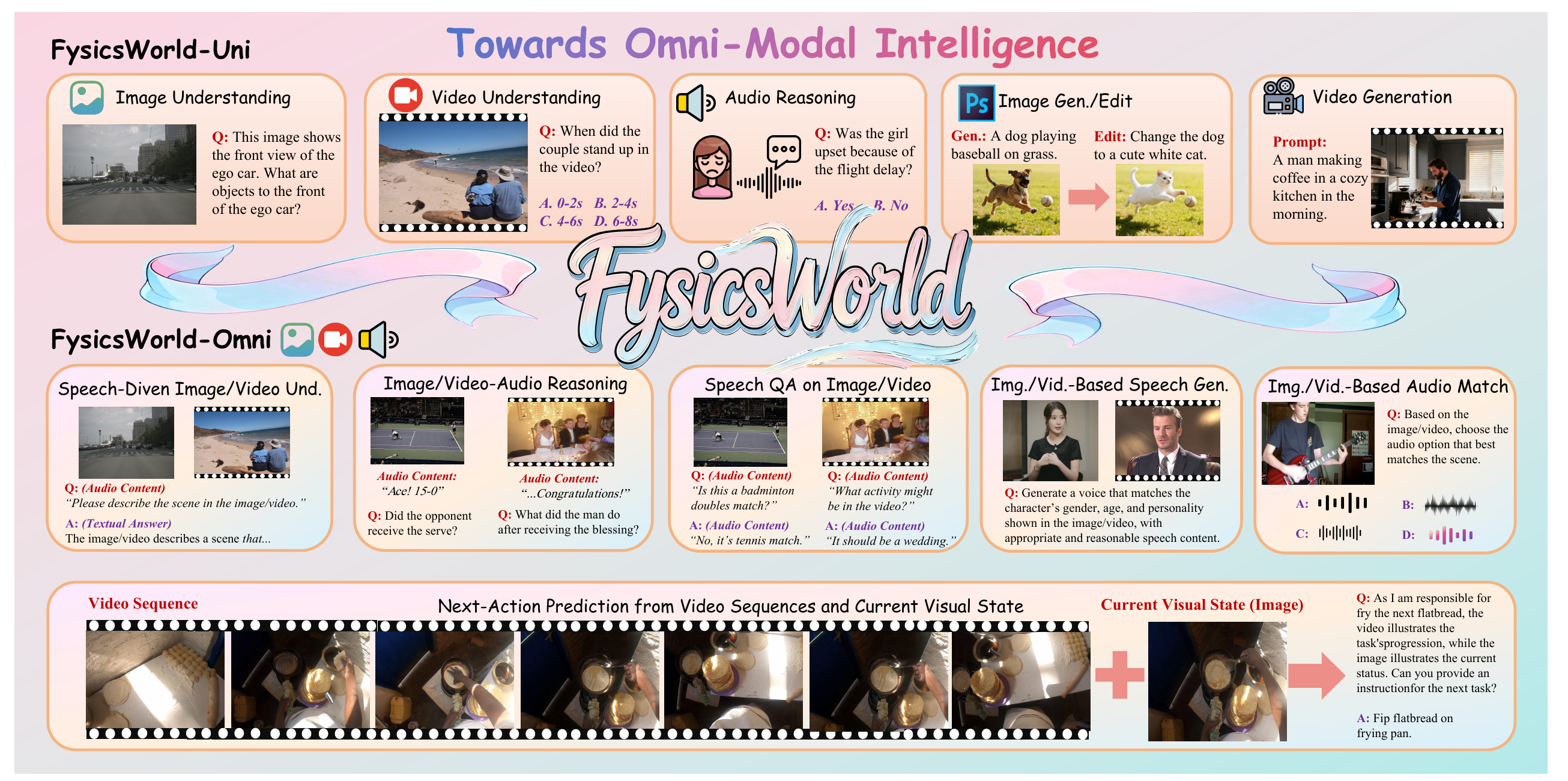}
    \caption{Examples of 16 comprehensive tasks in \textbf{\textit{FysicsWorld}}, categorized into \textbf{\textit{FysicsWorld-Uni}} and \textbf{\textit{FysicsWorld-Omni}}, supporting bidirectional I/O across image, video, audio, and text, enabling any-to-any evaluation across understanding, generation, and reasoning.}
    \label{fig-teaser}
\end{figure}

Multimodal large language models (MLLMs)~\cite{achiam2023gpt, bai2025qwen25vl, liu2023llava} are undergoing a rapid paradigm shift. Beyond extending the linguistic capabilities of traditional LLMs, the emerging research frontier aims to develop omni-modal large language models (OmniLLMs)—unified architectures capable of jointly processing and generating information across text, vision, audio, and potentially additional sensory modalities. Such architectures aspire not merely to perceive omni-modal content, but to integrate visual understanding and generation within a single model, enabling synergistic interactions among modalities. This shift is motivated by the complexity of the physical world: real-world intelligence hinges on the ability to integrate information richer than text alone—visual cues, auditory signals, spatial dynamics—and to respond to subtle multimodal interactions that govern perception and action.

Accompanying the rise of OmniLLMs~\cite{xu2025qwen3omni,xu2025qwen25omni} and unified understanding–generation models~\cite{deng2025bagel,chen2025blip3onext}, a wide array of benchmarks~\cite{zhou2025dailyomni, shi2025realunify, jiang2025danmakutppbench} has emerged to probe multimodal abilities. Despite this proliferation, existing benchmarks exhibit several fundamental limitations.
First, modality coverage remains incomplete. Many benchmarks only combine images or videos with audio in limited ways, falling to capture the breadth of full-modality interactions present in real-world environments.
Second, current strategies of omni-modal data construction largely rely on shallow modality concatenation, neglecting intrinsic cross-modal dependencies. As a result, tasks often permit single-modality shortcuts, allowing models to succeed without genuine multimodal fusion, alignment, or reasoning.
Third, nearly all existing omni-modal benchmarks remain text-centric, evaluating cross-modal understanding but not multimodal generation, and almost entirely omitting speech-driven interaction, which is the interface for real-world communication. These constraints hinder systematic assessment of the next generation of full-modality intelligent systems.

\begin{table}[]
\centering
\caption{\textbf{Comparison of Omni-modal Datasets and Benchmarks.} I, V, A, T represent image, video, audio, and text, respectively. CE and OE denote closed-ended and open-ended questions, while MCQ and GEN stand for multiple-choice questions and generative questions, respectively. Designed for real-world scenarios, \textbf{\textit{FysicsWorld}} exhibits significant data diversity and stands as the only benchmark supporting bidirectional full-modality I/O, characterized by strong interdependence and complementarity among modalities.}
\label{tab-related}
\renewcommand{\arraystretch}{1.2}% 
\resizebox{\textwidth}{!}{%
\begin{tabular}{lcccccccc}
\hline
\textbf{Benchmark/ Dataset} & \textbf{Modality} & \textbf{\begin{tabular}[c]{@{}c@{}}Multimodal \\ Output\end{tabular}} & \textbf{Tasks}    & \textbf{Questions} & \textbf{Question Type}                                      & \textbf{Open-Domain} & \textbf{\begin{tabular}[c]{@{}c@{}}Real-World \\ Scenarios\end{tabular}} & \textbf{\begin{tabular}[c]{@{}c@{}}Modality\\ Correlations\end{tabular}} \\ \hline 
\rowcolor{diy_pink} \multicolumn{9}{c}{\textit{\textbf{Uni-Modal Benchmarks}}}                                                                                                                                                                                                                                                                                                \\
MME~\cite{fu2025mme}                         & I+T               & \textcolor{red}{\ding{55}}                                                                     & 1                 & 2,194               & CE                                                          & \textcolor{green}{\ding{51}}                    & \textcolor{red}{\ding{55}}                   & \textcolor{red}{\ding{55}}                                                                        \\
MVBench~\cite{li2024mvbench}                     & V+T               & \textcolor{red}{\ding{55}}                                                                     & 9                & 4,000               & MCQ                                                         & \textcolor{green}{\ding{51}}                    & \textcolor{red}{\ding{55}}                   & \textcolor{red}{\ding{55}}                                                                        \\
MMAU~\cite{sakshi2024mmau}                        & A+T               & \textcolor{red}{\ding{55}}                                                                     & 2                & 10,000              & MCQ                                                         & \textcolor{green}{\ding{51}}              & \textcolor{red}{\ding{55}}                   & \textcolor{red}{\ding{55}}                                                                        \\ \hline
\rowcolor{diy_pink} \multicolumn{9}{c}{\textit{\textbf{Omni-Modal Benchmarks}}}                                                                                                                                                                                                                                                                                               \\
OmniBench~\cite{li2024omnibench}                   & I+A+T             & \textcolor{red}{\ding{55}}                                                                     & 8                 & 1,142               & MCQ                                                         & \textcolor{green}{\ding{51}}                    & \textcolor{red}{\ding{55}}                   & A-I                                                                      \\
OmniMMI~\cite{wang2025omnimmi}                     & V+A+T             & \textcolor{red}{\ding{55}}                                                                     & 2                 & 2,290               & MCQ                                                         & \textcolor{green}{\ding{51}}                    & \textcolor{green}{\ding{51}}                   & A-V                                                                      \\
Daily-Omni~\cite{zhou2025dailyomni}                  & V+A+T             & \textcolor{red}{\ding{55}}                                                                     & 2   & 1,197               & MCQ                                                         & \textcolor{green}{\ding{51}}               & \textcolor{green}{\ding{51}}                   & A-V                                                                      \\
HumanSense~\cite{qin2025humansense}                  & V+A+T             & \textcolor{red}{\ding{55}}                                                                     & 4  & 3,882               & OE, MCQ                                                     & \textcolor{red}{\ding{55}}                     & \textcolor{red}{\ding{55}}                   & A-V                                                                      \\
OmniVideoBench~\cite{li2025omnivideobench}              & V+A+T             & \textcolor{red}{\ding{55}}                                                                     & 13                & 1,000               & MCQ                                                         & \textcolor{green}{\ding{51}}                & \textcolor{red}{\ding{55}}                   & A-V                                                                      \\
LongVALE~\cite{geng2025longvale}                    & V+A+T             & \textcolor{red}{\ding{55}}                                                                     & 2                 & 8,411               & OE                                                          & \textcolor{green}{\ding{51}}                    & \textcolor{red}{\ding{55}}                   & A-V                                                                      \\
AVHBench~\cite{sung2024avhbench}                    & V+A+T             & \textcolor{red}{\ding{55}}                                                                     & 4                 & 5,186               & CE, OE                                                       & \textcolor{red}{\ding{55}}                    & \textcolor{red}{\ding{55}}                   & A-V                                                                      \\
WorldSense~\cite{hong2025worldsense}                 & V+A+T             & \textcolor{red}{\ding{55}}                                                                     & 8  & 3,172               & MCQ                                                         & \textcolor{green}{\ding{51}}                & \textcolor{green}{\ding{51}}                   & A-V                                                                      \\
AV-Odyssey Bench~\cite{gong2024avodyssey}            & I+V+A+T           & \textcolor{red}{\ding{55}}                                                                     & 26                & 4,555               & MCQ                                                         & \textcolor{green}{\ding{51}}               & \textcolor{red}{\ding{55}}                   & A-V, A-I                                                                 \\ \hline
\textbf{\textit{FysicsWorld}} (Ours)           & I+V+A+T           & \textcolor{green}{\ding{51}}(I+V+A+T)                                                            & 16 & 3,268               & \begin{tabular}[c]{@{}c@{}}OE, CE, \\ MCQ, GEN\end{tabular} & \textcolor{green}{\ding{51}}               & \textcolor{green}{\ding{51}}                   & A-V, A-I, A-V-I                                                          \\ \hline
\end{tabular}
}
\renewcommand{\arraystretch}{1.0}%

\end{table}

To address these gaps, we introduce \textbf{\textit{FysicsWorld}}, the first unified full-modality benchmark that supports bidirectional input–output across image, video, audio, and text, with carefully curated cross-modal dependencies and complementarities. \textbf{\textit{FysicsWorld}} enables comprehensive any-to-any evaluation across understanding, generation, and reasoning, providing a unified platform for examining how models perceive, align, fuse, and generate information.

Our benchmark consists of two complementary subsets: \textbf{\textit{FysicsWorld-Uni}}, which focuses on uni-modal understanding and generation, and \textbf{\textit{FysicsWorld-Omni}}, which targets omni-modal interaction and fusion-dependent cross-modal reasoning. Together, \textbf{\textit{FysicsWorld}} contains 3,268 curated samples, spanning 16 task categories and over 226 fine-grained sub-tasks, covering 179 open-domain topics. Representative examples are illustrated in Figure~\ref{fig-teaser}, and a detailed taxonomy is provided in Table~\ref{tab-task}.

We also propose a construction method for omni-modal data, which is named \textbf{C}ross-\textbf{M}odal \textbf{C}omplementarity \textbf{S}creening (CMCS) strategy, integrated within a systematic construction framework for generating high-quality omni-modal data for speech-driven interaction and fusion-dependent reasoning. CMCS ensures that the resulting tasks maintain strong cross-modal coupling, preventing single-modality shortcuts and enforcing true multimodal reasoning. Collectively, \textbf{\textit{FysicsWorld}} exhibits multi-dimensional, multi-modal, multi-task, multi-source, multi-domain, multi-type, multi-target, and multi-assurance characteristics, as detailed in Section~\ref{sec-3-1}.

\begin{figure}[t]
    \centering
    \includegraphics[width=\textwidth]{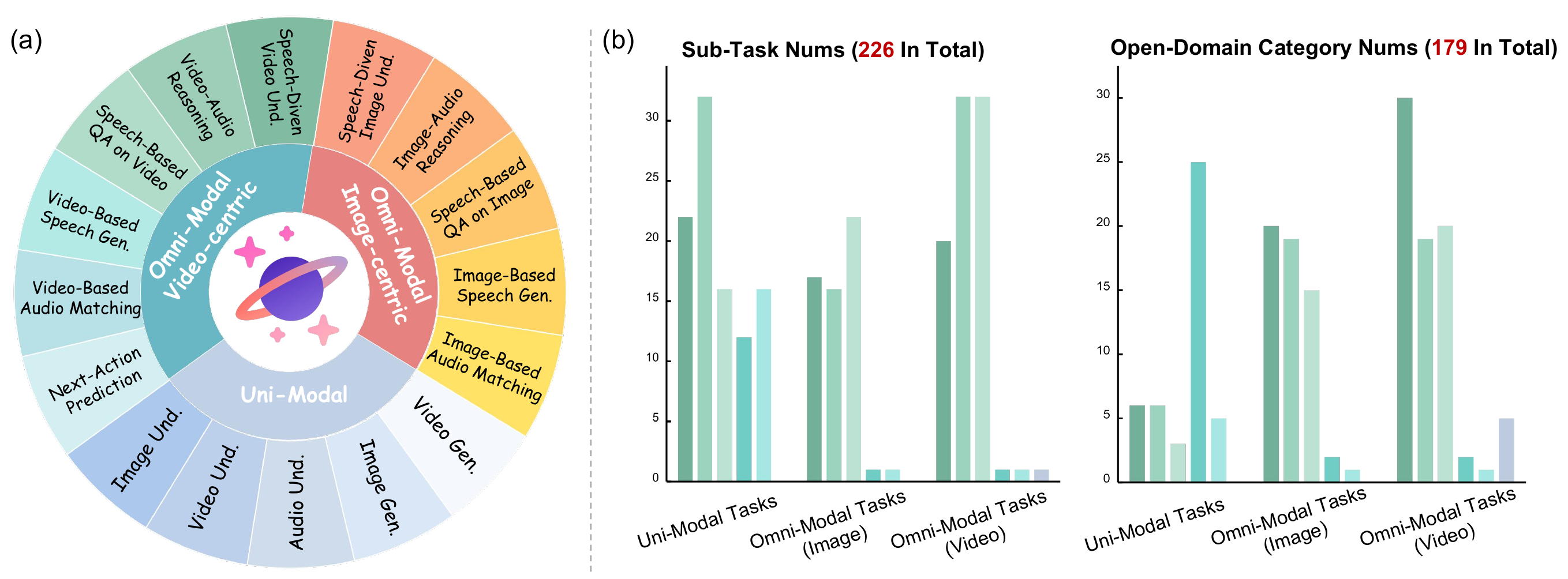}
    \caption{\textbf{Data statistics of \textit{FysicsWorld}.}
    (a) FysicsWorld encompasses full-modality coverage across 16 major task types, spanning uni-modal settings, image-centric omni-modal tasks, and video-centric omni-modal tasks.
    (b) Across these task types, \textbf{\textit{FysicsWorld}} provides a total of 226 fine-grained sub-tasks, including complex real-world scenarios such as attribute and motion recognition for multiple vehicles in autonomous driving. In addition, \textbf{\textit{FysicsWorld}} covers 179 fine-grained open-domain categories, broadly spanning daily life and diverse real-world environments.}
    \label{fig-task}
\end{figure}

Through extensive evaluation of state-of-the-art models (including OmniLLMs, MLLMs, unified understanding–generation, and modality-specialized models), our benchmark reveals fundamental performance disparities and capability boundaries across full-modality tasks. This comprehensive analysis provides strong baselines and a unified foundation for advancing research in omni-modal learning and general-purpose multimodal intelligence.
Our main contributions are summarized as follows:
\begin{itemize}
    \item We present \textbf{\textit{FysicsWorld}}, the first unified benchmark with bidirectional full-modal I/O, supporting any-to-any evaluation across understanding, generation, and reasoning.
    \item We introduce the CMCS strategy and a systematic data construction framework for realistic spoken interaction and fusion-dependent cross-modal reasoning.
    \item We evaluate over 30 OmniLLMs, MLLMs, unified understanding–generation models, and modality-specialized models, revealing key limitations across architectures and establishing strong baselines for future development of unified omni-modal models.
\end{itemize}

%%%%%% Related Works %%%%%%
\section{Related Work}
\label{sec:related}

\subsection{Datasets for Uni-Modal Tasks}
The recent rapid development of vision-language models (VLMs)~\cite{liu2023llava, bai2025qwen25vl, zhu2025internvl3} and audio-language models (ALMs)~\cite{chu2024qwen2audio, tang2023salmonn} has led to the emergence of numerous benchmarks designed to evaluate their multimodal perception and generation capabilities.
Given considerations in multimodal learning~\cite{yang2024asynchronous,yang2024towards,yang2022disentangled,yang2025improvingmsa,yang2024pediatricsgpt,yang2025medaide,liu2025reinforcement,lin2025sail,jiang2025satiredecoder}, we outline the following for each modality.

\vspace{0.3pt}\noindent\textbf{Image Modality.}\hspace{1ex}
Existing works target distinct aspects of visual understanding and reasoning. MMMU~\cite{yue2024mmmu} focuses on university-level subject knowledge; MME~\cite{fu2025mme} and MMBench~\cite{liu2024mmbench} measure general-purpose visual understanding capabilities; OCRBench~\cite{liu2024ocrbench}, MathVista~\cite{lu2023mathvista}, and HallusionBench~\cite{guan2024hallusionbench} evaluate OCR competence, mathematical reasoning, and hallucination resistance, respectively; WISE~\cite{niu2025wise} and GEdit-Bench~\cite{step1x2025geditbench} provide systematic evaluations for image generation and controllable editing.

\vspace{0.3pt}\noindent\textbf{Vision Modality.}\hspace{1ex}
Related works now address temporal-semantic reasoning and generation. MVBench~\cite{li2024mvbench} provides 20 challenging understanding tasks, LongVideoBench~\cite{wu2024longvideobench} targets hour-long video comprehension, and VBench~\cite{huang2024vbench} assesses generation quality across fidelity, aesthetics, motion coherence, and stability.

\vspace{0.3pt}\noindent\textbf{Audio Modality.}\hspace{1ex}
Beyond traditional automatic speech recognition (ASR) and text-to-speech (TTS) tasks, several comprehensive evaluation suites have been proposed. MMAU~\cite{sakshi2024mmau} measures comprehension of speech, audio, and music; MMAR~\cite{ma2025mmar} extends this to the mixture of audio types reasoning; and MMSU~\cite{wang2025mmsu} probes semantic content, paralinguistic cues (\textit{e.g.,} emotion, tempo, pitch), and phonological structures embedded in speech.

Despite their breadth, these uni-modal benchmarks differ widely in task design, coverage, and difficulty. Their heterogeneous objectives and modality-specific focus complicate unified evaluation and hinder systematic analysis of emerging full-modality architectures.

\begin{table}[t]
\begingroup
\renewcommand{\arraystretch}{1.2} % 1.2x row height
\caption{\textbf{Detailed Taxonomy of the \textit{FysicsWorld} Benchmark.} The table outlines the 16 primary tasks, categorized into \textbf{\textit{FysicsWorld}}-Uni and \textbf{\textit{FysicsWorld}}-Omni (image-centric and video-centric). For each task, it specifies the task definition, Input/Output (I/O) format, question type, evaluation metric, and data source. For Task1-4, “(I)+T→I” indicates an optional image input: “T→I” denotes image generation, while “I+T→I” denotes text-guided image editing.
}
\label{tab-task}
\resizebox{\textwidth}{!}{%
\begin{tabular}{cccccc}
\hline
        & \textbf{Task Definition}                                                                                       & \textbf{I/O} & \textbf{Question Type} & \textbf{Metric}   & \textbf{Source} \\ \hline
\rowcolor{diy_pink} \multicolumn{6}{c}{\textit{\textbf{FysicsWorld-Uni}}}                                                                                                                            \\
Task1-1 & Image Understanding                                                                                            & I+T→T        & OE                     & ACC, BERTScore    & public          \\
Task1-2 & Video Understanding                                                                                            & V+T→T        & CE, MCQ                & ACC               & public          \\
Task1-3 & Audio Reasoning                                                                                                & A+T→T        & MCQ                    & ACC               & public          \\
Task1-4 & Image Generation                                                                                               & (I)+T→I      & GEN                    & WIScore, VIEScore & public          \\
Task1-5 & Video Generation                                                                                               & T→V          & GEN                    & VQ                & public          \\ \hline
\rowcolor{diy_pink} \multicolumn{6}{c}{\textit{\textbf{FysicsWorld-Omni (image-centric)}}}                                                                                                                                   \\
Task2-1 & Speech-Driven Image Understanding                                                                              & I+A→T        & MCQ                    & ACC               & synthetic       \\
Task2-2 & Image–Audio Contextual Reasoning                                                                               & I+A+T→T      & MCQ                    & ACC               & synthetic       \\
Task2-3 & Speech-Based QA on Image Content                                                                               & I+A→A        & GEN                    & ASR-BLEU, SIM     & synthetic       \\
Task2-4 & Speech Generation from Person in Image                                                                         & I+T→A        & GEN                    & IC, NLQ           & synthetic       \\
Task2-5 & Audio Matching from Image Context                                                                              & I+A+T→T      & MCQ                    & ACC               & synthetic       \\ \hline
\rowcolor{diy_pink} \multicolumn{6}{c}{\textit{\textbf{FysicsWorld-Omni (video-centric)}}}                                                                                                                                   \\
Task3-1 & Speech-Driven Video Understanding                                                                              & V+A→T        & MCQ                    & ACC               & synthetic       \\
Task3-2 & Video–Audio Contextual Reasoning                                                                               & V+A+T→T      & MCQ                    & ACC               & synthetic       \\
Task3-3 & Speech-Based QA on Video Content                                                                               & V+A→A        & GEN                    & ASR-BLEU, SIM     & synthetic       \\
Task3-4 & Speech Generation from Person in Video                                                                         & V+T→A        & GEN                    & IC, NLQ           & synthetic       \\
Task3-5 & Audio Matching from Video Context                                                                              & V+A+T→T      & MCQ                    & ACC               & synthetic       \\
Task3-6 & \begin{tabular}[c]{@{}c@{}}Next-Action Prediction from Video\\ Sequences and Current Visual State\end{tabular} & V+I+T→T      & MCQ                    & ACC               & synthetic       \\ \hline
\end{tabular}
}
\renewcommand{\arraystretch}{1.0}
\endgroup

\end{table}

\subsection{Datasets for Omni-Modal Tasks}
With the rise of omni-modal architectures, the demand for benchmarks capable of evaluating cross-modal integration, alignment, reasoning, and generation has become increasingly urgent. As summarized in Table~\ref{tab-related}, several recent efforts attempt to move in this direction.
OmniBench~\cite{li2024omnibench} and AV-Odyssey~\cite{gong2024avodyssey} assess joint image-audio recognition. HumanSense~\cite{qin2025humansense} and AVHBench~\cite{sung2024avhbench} are domain-specific with monotonous queries. Daily-Omni~\cite{zhou2025dailyomni}, WorldSense~\cite{hong2025worldsense}, and OmniMMI~\cite{wang2025omnimmi} target real-world scenarios but are constrained by text-centric reasoning and limited multimodal interaction.

Among existing efforts, most datasets suffer from incomplete modality coverage and exhibit weak modality correlations, relying primarily on shallow modality concatenation and thereby preventing a reliable assessment of whether models truly perform multimodal fusion. Furthermore, none of the existing benchmarks support bidirectional, full-modality input–output, lacking cross-modal generation and interaction. To address these, \textbf{\textit{FysicsWorld}} is the first unified full-modality benchmark, with carefully curated modality dependencies that ensure strong complementarity rather than redundancy. This design enables comprehensive any-to-any evaluation across understanding, generation, and reasoning, paving the way for systematic evaluation of next-generation OmniLLMs.

%%%%%% Method %%%%%%
\section{\textbf{\textit{FysicsWorld}}}
\subsection{Overview of \textbf{\textit{FysicsWorld}}}

\begin{figure}[t]
    \centering
    \includegraphics[width=\textwidth]{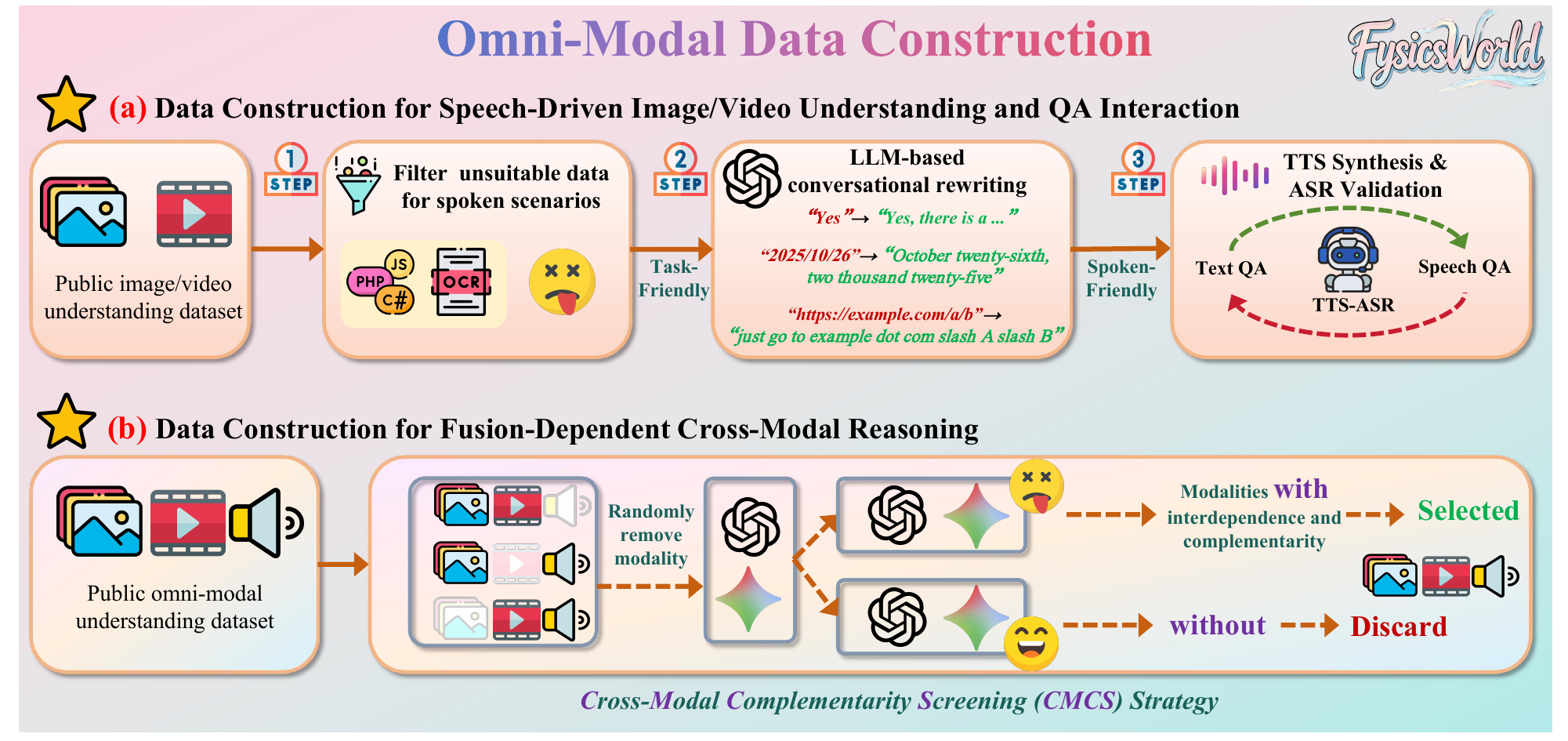}
    \caption{\textbf{Construction framework for }\textit{\textbf{FysicsWorld-Omni}}\textbf{ data.} The framework illustrates two key pipelines: (a) Data Construction for Speech-Driven Understanding and QA Interaction and (b) the Cross-Modal Complementarity Screening (CMCS) strategy for generating Fusion-Dependent Cross-Modal Reasoning tasks.}
    
    \label{fig-pipeline}
\end{figure}

\begin{figure}[t]
    \centering

    \newsavebox{\tempimagebox}
    \sbox{\tempimagebox}{\includegraphics[width=\textwidth]{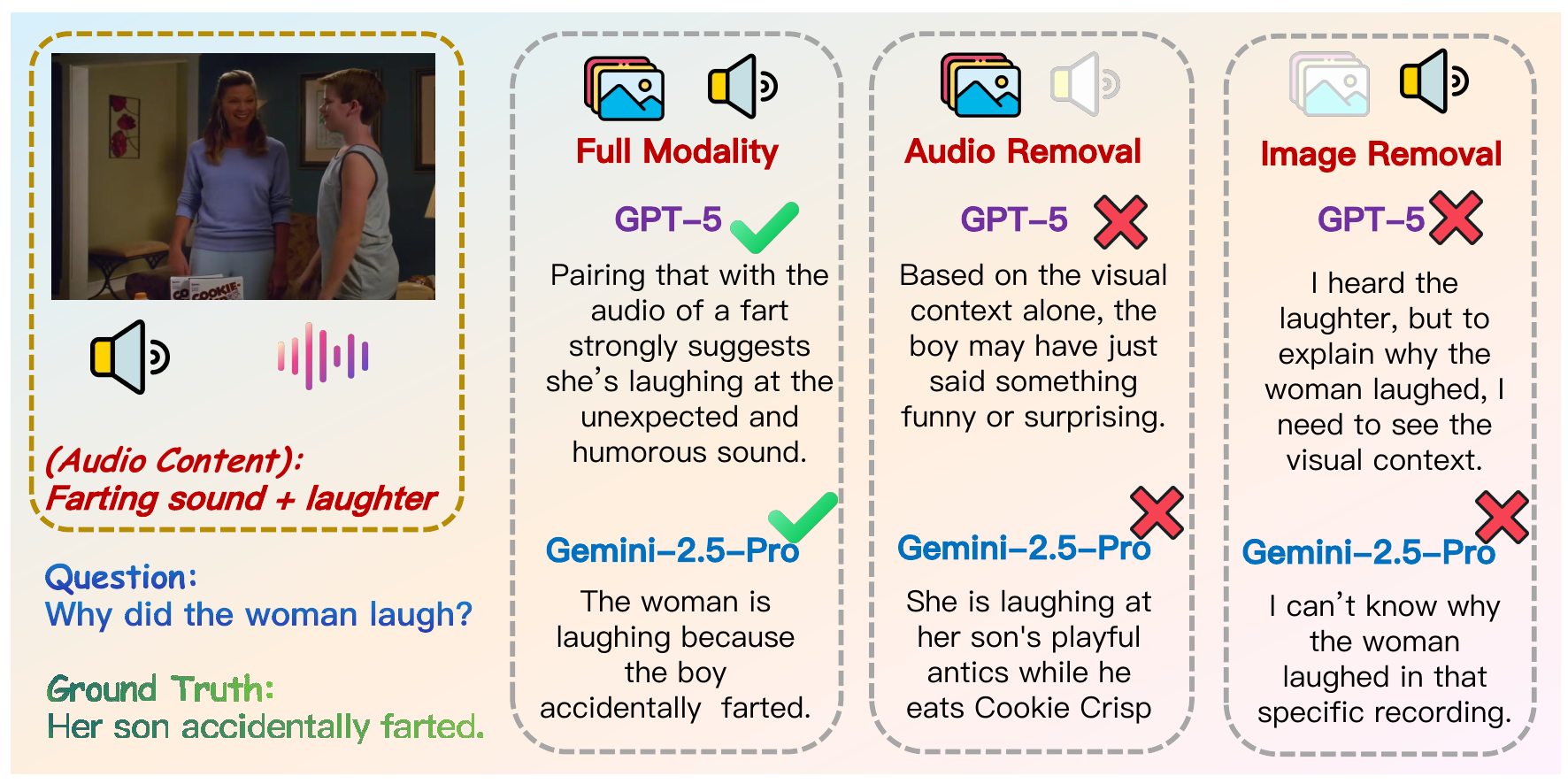}}

    \includegraphics[width=\textwidth, height=0.9\ht\tempimagebox]{Figure/supplement-fig/CMCS-1.pdf}

    \caption{\textbf{Example of samples retained by the CMCS strategy.}This instance exhibits strong cross-modal dependency: ablating any single modality leads to substantial degradation in MLLMs' performance. The preserved samples by our CMCS strategy require genuine multimodal fusion—rather than unimodal shortcuts—to be solved correctly.}
    \label{fig-sup-cmcs-1}
\end{figure}

\begin{figure}[t]
    \centering

    % \newsavebox{\tempimagebox}
    \sbox{\tempimagebox}{\includegraphics[width=\textwidth]{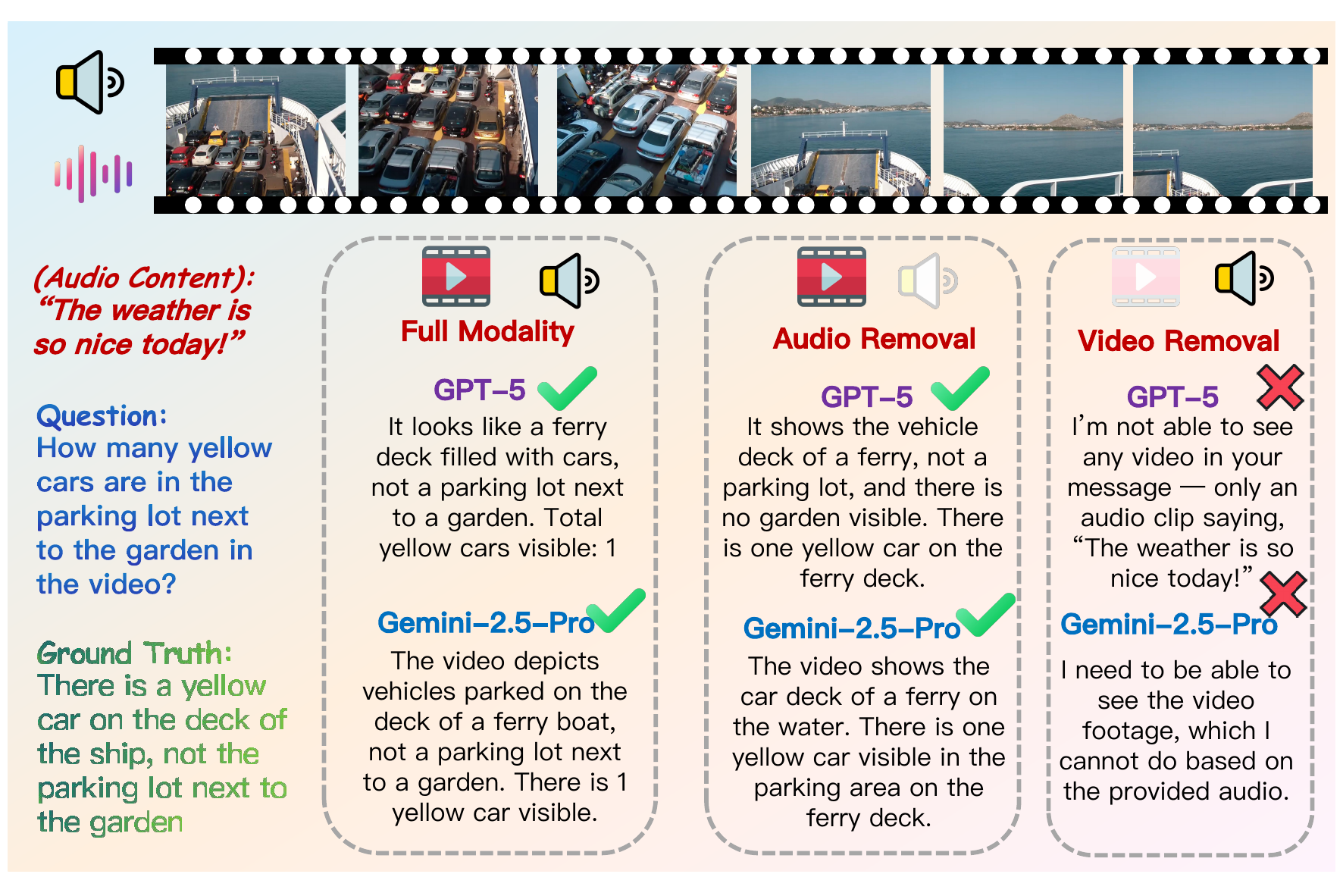}}

    \includegraphics[width=\textwidth, height=0.9\ht\tempimagebox]{Figure/supplement-fig/CMCS-2.pdf}

    \caption{\textbf{Example of samples filtered out by the CMCS strategy.} This case can be correctly answered after audio modality ablation, indicating weak or redundant modality coupling. It can be solved using single-modality cues (e.g., video-only reasoning), making it unsuitable for evaluating fusion-dependent cross-modal reasoning.}
    \label{fig-sup-cmcs-2}
\end{figure}

To bridge the gaps left by existing omni-modal benchmarks, probe the capability boundaries of OmniLLMs, MLLMs, unified understanding-generation models, and modality-specialized models, we introduce \textbf{\textit{FysicsWorld}}, the first unified full-modality benchmark enabling bidirectional input–output across image, video, audio, and text, supporting comprehensive any-to-any evaluation across understanding, generation, and reasoning.

\vspace{0.3pt}\noindent\textbf{Data Stastics.}\hspace{1ex}
\label{sec-3-1}
As shown in Figure~\ref{fig-task}, our benchmark is characterized by eight “\textit{multi}” properties, reflecting its comprehensive coverage, diversity, and robustness, namely:
\textit{multi-dimensional} (understanding, generation, reasoning, voice interaction), \textit{multi-modal} (text, image, video, audio as both inputs and outputs), \textit{multi-task} (16 primary tasks, 226 sub-tasks), \textit{multi-source} (3,268 samples from 40+ public datasets and curated web data), \textit{multi-domain} (179 open-domain categories), \textit{multi-type} (closed-ended, open-ended, multiple-choice question, and image/video/audio generation), \textit{multi-target} (evaluates OmniLLMs, MLLMs, modality-specific models, unified understanding–generation models), and \textit{multi-assurance} (multi-stage quality control strategies).

\vspace{0.3pt}\noindent\textbf{Task Taxonomy.}\hspace{1ex}
Our benchmark consists of 16 comprehensive tasks, as illustrated in Figure~\ref{fig-teaser}, which can be divided into two subsets: (1) \textbf{\textit{FysicsWorld-Uni}}, comprising 5 foundational uni-modal tasks, serving to evaluate various multimodal models on their foundational understanding and generation capabilities. (2) \textbf{\textit{FysicsWorld-Omni}}, encompassing 11 omni-modal tasks, which are designed to explore the performance of OmniLLMs and MLLMs under real-world, full-modality intelligent scenarios. The detailed taxonomy of our benchmark is presented in Table~\ref{tab-task}. In the following sections, we describe the construction pipeline and design principles behind \textbf{\textit{FysicsWorld}} in detail.

\subsection{Construction of \textbf{\textit{FysicsWorld-Uni}}}
Despite the proliferation of open-source uni-modal benchmarks~\cite{fu2025mme, wu2024longvideobench}, their objectives, task coverage, and design philosophies vary substantially, leading to a fragmented evaluation landscape that insufficiently reflects the multifaceted nature of multimodal reasoning in open-domain scenes. Besides, few existing resources consider real-world robustness or semantic comprehensiveness, leaving gaps in evaluating high-level reasoning, generalization, and multimodal alignment under complex natural inputs.

To address these deficiencies, \textbf{\textit{FysicsWorld-Uni}} is constructed through a comprehensive multi-source synthesis and refinement pipeline.
We curate data from over 40 uni-modal datasets, selectively integrating complementary, high-quality instances that capture diverse reasoning dimensions, perceptual challenges, and content domains. Low-quality annotations are manually corrected via a human-LLM collaborative review, and we expand real-world visual and audio coverage to better reflect open-environment interactive scenarios. Details for different tasks are as follows:

\vspace{0.3pt}\noindent\textbf{Image Understanding.}\hspace{1ex}
Evaluation spans general VQA, university-level reasoning, math, OCR/charts, and hallucination, using data from MME~\cite{fu2025mme}, MMMU~\cite{yue2024mmmu}, MathVista~\cite{lu2023mathvista}, MMVP~\cite{tong2024mmvp}, HallusionBench~\cite{guan2024hallusionbench}, and real-world sets MME-RealWorld~\cite{zhang2024mmerealworld} and SEED-Bench-H~\cite{li2023seedbench}.

\vspace{0.3pt}\noindent\textbf{Video Understanding.}\hspace{1ex}
Built from the Video-MME~\cite{fu2025videomme} dataset and MVBench~\cite{li2024mvbench}, with the streaming videos from OmniMMI~\cite{wang2025omnimmi} to stress real-world temporal reasoning. Multiple-choice candidate pools are refined to reduce ambiguity and enforce discriminative distractors, strengthening tests of temporal, causal, and spatial comprehension.

\vspace{0.3pt}\noindent\textbf{Audio Reasoning.}\hspace{1ex}
Integrated from four complementary benchmarks, MMAU~\cite{sakshi2024mmau}, MMAR~\cite{ma2025mmar}, MMSU~\cite{wang2025mmsu}, and AIR-Bench~\cite{yang2024airbench}, to evaluate speech, sound, and music across more than 20 sub-tasks, probing both perceptual recognition and semantic reasoning. 

\vspace{0.3pt}\noindent\textbf{Image/Video Generation.}\hspace{1ex}
Integrates WISE~\cite{niu2025wise}, GEdit-Bench~\cite{step1x2025geditbench}, VBench~\cite{huang2024vbench}, and Video-Bench~\cite{han2025videobench} to assess instruction-conditioned synthesis, controllable editing, and temporal consistency. This unification expands the breadth and thematic diversity, ensuring a more comprehensive and fine-grained assessment of visual creativity, fidelity, and alignment with complex prompts.

\subsection{Construction of \textbf{\textit{FysicsWorld-Omni}}}
To overcome the limitations identified in prior omni-modal benchmarks, we introduce \textbf{\textit{FysicsWorld-Omni}}. While most existing datasets limit multimodal evaluation to text-centric reasoning patterns, \textbf{\textit{FysicsWorld-Omni}} extends the paradigm toward real-world voice-interactive understanding and generation. This subset encompasses 11 tasks organized into three principal categories: (i) speech-driven image \& video understanding and QA interaction, (ii) fusion-dependent cross-modal reasoning, and (iii) cross-modal audio generation. These tasks jointly explore how OmniLLMs and MLLMs operate when understanding, reasoning, and generation must interact seamlessly. The omni-modal data construction framework is illustrated in Figure~\ref{fig-pipeline}.

\vspace{0.3pt}\noindent\textbf{Speech-Driven Image/Video Understanding and QA Interaction.}\hspace{1ex}
To support natural, multimodal communication, we develop a speech-grounded multimodal data construction pipeline that ensures both linguistic fluency and semantic fidelity in voice-based interactions.
Starting from high-quality public image/video understanding datasets, we first filter tasks unsuitable for speech scenarios (\textit{e.g.,} OCR or code-reasoning tasks). Then LLM-based conversational rewriting enhances textual QA pairs, expanding terse answers, reformulating numerals and symbols into spoken-friendly forms, and converting incomplete or formal phrasing into natural, oral expressions. The rewriting yields dialogue-style instructions more aligned with real spoken interaction.
Each rewritten sample is synthesized into audio using TTS with 20 randomly selected voices, differing in tone, pitch, and timbre, to enrich diversity and simulate authentic human variability.
To ensure semantic alignment between spoken and textual content, each synthesized voice is validated by \textit{AlignScore} via ASR:
\vspace{-3pt}
\begin{equation}
\text{\textit{AlignScore}} = 1 - \mathrm{WER}\big(\mathrm{ASR}(\mathrm{TTS}(y)),\, y\big),
\end{equation}
where $y$ denotes the original text and $WER()$ is the Word Error Rate between the ASR text and the original text. Lower word error rate indicates better agreement, so $AlignScore \in [0,1]$ increases as the synthesized speech becomes more semantically consistent with the original text. Samples falling below the $AlignScore$ threshold are re-synthesized or discarded. The resulting corpus comprises both speech-driven visual instruction tasks and spoken QA tasks. Together, they provide a rigorous platform for assessing the capabilities of speech-driven cross-modal interaction.

\vspace{0.3pt}\noindent\textbf{Fusion-Dependent Cross-Modal Reasoning.}\hspace{1ex}
Existing omni-modal datasets typically combine two or three modalities with weak interdependence, allowing models to solve problems using only a single modality without information fusion. Such simplifications obscure whether success stems from cross-modal reasoning or superficial understanding.

To ensure that every modality contributes essential, non-redundant information, we introduce a principled mechanism termed the \textbf{C}ross-\textbf{M}odal \textbf{C}omplementarity \textbf{S}creening (CMCS) strategy, as illustrated in Figure~\ref{fig-pipeline}.
Conceptually, CMCS operates as a fusion-dependency discovery process. We employ advanced MLLMs, GPT-5~\cite{openai_gpt5_system_card} and Gemini-2.5-Pro~\cite{comanici2025gemini25}, to first evaluate full multimodal inputs. Subsequently, each candidate sample undergoes a selective modality ablation process, where a single modality is randomly removed from the input stream.
By comparing model accuracy on the complete versus ablated inputs, we measure the performance degradation attributable to the missing modality. Samples yielding substantial degradation across the MLLMs are retained as fusion-dependent cases, meaning the task cannot be solved without integrating multiple complementary modalities.
This cross-modal complementarity filtering ensures that all selected tasks require authentic multimodal reasoning, rather than relying on isolated cues.
The resulting subset comprises the fusion-dependent cross-modal reasoning tasks in \textbf{\textit{FysicsWorld-Omni}}, ensuring that each retained instance requires cooperative inference across vision, audio, and language, thereby minimizing modality redundancy and bridging the semantic gap inherent to multimodal learning.

We present more detailed visualizations and comparative results for the CMCS-based data selection process. As illustrated in Figure~\ref{fig-sup-cmcs-1}, it shows a representative example that passes CMCS screening. Regardless of whether image or audio modal information is removed from the multimodal inference sample, advanced MLLMs, such as GPT-5 and Gemini-2.5-Pro, cannot resolve the issue, indicating significant information complementarity and coupling between the modalities in this data sample. The samples retained by the CMCS strategy demonstrate strong cross-modal coupling, where removing any single modality leads to a notable decline in expert model responses and indicates that correct reasoning requires integrating complementary evidence across vision, audio, and linguistic cues. Conversely, Figure~\ref{fig-sup-cmcs-2} presents an example that is filtered out by CMCS. These samples can be answered correctly even after modality ablation, revealing redundant or weak cross-modal dependencies. Retaining such cases would permit unimodal shortcuts and undermine the goal of evaluating genuine multimodal fusion. These analyses demonstrate the effectiveness of CMCS in identifying samples requiring genuine multimodal fusion.

\vspace{-5pt}

\subsection{Quality Assurance and Ethical Considerations}
To ensure quality and ethical compliance, we adopt a multi-assurance strategy. During data collection, we rely on publicly available, high-quality datasets and perform human screening to remove incomplete, ambiguous, or low-fidelity samples. We then apply sensitive content filtering, using an open-source stop-word list covering advertisements, profanity, drugs, gambling, politics, pornography, violence, phishing URLs, and other high-risk categories.
During omni-modal synthesis, we use LLM-based conversational rewriting to obtain spoken-friendly text, TTS–ASR consistency checking to enforce audio–text alignment, and the CMCS strategy to verify fusion dependence. All data comes from licensed public sources and excludes personally identifiable information; synthetic speech does not imitate identifiable voices.
%%%%%% Experiments %%%%%%
\section{Experiments}
In this section, we present a comprehensive evaluation of state-of-the-art OmniLLMs, MLLMs, modality-specific models, and unified understanding–generation models on \textbf{\textit{FysicsWorld}}, revealing the coexistence of opportunities and challenges in advancing future full-modality modeling, perception, understanding, and generation.

\begin{table}[t]
\caption{\textbf{Performance Comparison of OmniLLMs and MLLMs on Image-centric Tasks.} The table details model performance on Task 1-1 (Image Understanding) and the omni-modal Tasks 2-1 to 2-5. Metrics for the speech generation task (Task 2-4) include identity consistency (IC) and natural language quality (NLQ). SIM represents speaker similarity. For all metrics, larger values indicate better performance.}
\begin{tabular}{lccccccccc}
\hline
\multirow{2}{*}{\textbf{Model}} & \multicolumn{2}{c}{\textbf{Task1-1}} & \textbf{Task2-1} & \textbf{Task2-2} & \multicolumn{2}{c}{\textbf{Task2-3}} & \multicolumn{2}{c}{\textbf{Task2-4}} & \textbf{Task2-5} \\
                                & ACC               & BERTScore        & ACC              & ACC              & BLEU              & SIM              & IC                & NLQ              & ACC              \\ \hline
\rowcolor{diy_pink}  \multicolumn{10}{c}{\textit{\textbf{OmniLLMs}}}                                                                                                                                                               \\
Qwen2.5-Omni-7B                 & 60.95             & 0.765            & 66.67            & 60.47            & 59.80             & 51.52            & 37.05             & 2.85             & 60.40            \\
Qwen3-Omni-30B-A3B              & \textbf{72.86}             & \textbf{0.809}            & \textbf{72.89}   & \textbf{66.05}            & \textbf{66.57}    & \textbf{61.35}   & \textbf{58.84}    & \textbf{3.69}    & \textbf{70.47}   \\
VITA-1.5                        & 52.86             & 0.718            & 55.56            & 55.81            & 52.30             & 43.81            & 28.91             & 2.00             & 54.36            \\
Stream-Omni                     & 53.33             & 0.748            & 58.22            & 65.12            & 52.82             & 50.74            & 40.15             & 2.13             & 54.36            \\
Ming-lite-Omni-1.5              & 60.95             & 0.761            & 67.56            & 62.33            & 61.15             & 53.62            & 52.70             & 3.65             & 60.40            \\
Baichuan-Omni-1.5               & 56.19             & 0.752            & 60.00            & 54.88            & 55.46             & 49.85            & 39.23             & 2.05             & 52.35            \\
MiniCPM-o 2.6                   & 57.62             & 0.742            & 60.00            & 57.21   & 54.70             & 45.23            & 32.50             & 2.18             & 53.02            \\ \hline
\rowcolor{diy_pink}  \multicolumn{10}{c}{\textit{\textbf{MLLMs}}}                                                                                                                                                                  \\
GPT-5                           & \textbf{75.71}    & 0.872            & 68.89            & \textbf{70.23}   & 68.13             & 58.62            & 55.48             & 3.85             & 70.47            \\
Gemini-2.5-Pro                  & 69.52             & \textbf{0.885}   & \textbf{73.78}   & 68.37            & -                 & -                & -                 & -                & \textbf{72.15}   \\ \hline
\end{tabular}
\vspace{-5pt}
\label{tab-exp-image}
\end{table}

\vspace{10pt}

\begin{table}[!tbp]
\caption{\textbf{Performance Comparison of OmniLLMs and MLLMs on Video-centric Tasks.} Abbreviations have the same meanings as those in Table~\ref{tab-exp-image}. For all metrics, larger values indicate better performance.}
\vspace{-10pt}
\begin{tabular}{lccccccccc}
\hline
\multirow{2}{*}{\textbf{Model}} & \multicolumn{1}{c}{\textbf{Task1-2}} & \textbf{Task3-1} & \textbf{Task3-2} & \multicolumn{2}{c}{\textbf{Task3-3}} & \multicolumn{2}{c}{\textbf{Task3-4}} & \textbf{Task3-5} & \textbf{Task3-6} \\
                                & \multicolumn{1}{c}{ACC}              & ACC              & ACC              & BLEU              & SIM              & IC                & NLQ              & ACC              & ACC              \\ \hline
\rowcolor{diy_pink}  \multicolumn{10}{c}{\textit{\textbf{OmniLLMs}}}                                                                                                                                                                                  \\
Qwen2.5-Omni-7B                 & 61.78                                & 45.89            & 38.64            & 50.38             & 57.26            & 39.05             & 2.98             & 60.07            & 51.27            \\
Qwen3-Omni-30B-A3B              & \textbf{67.56}                       & \textbf{53.62}   & \textbf{45.45}   & 57.73             & \textbf{63.14}   & 41.90             & \textbf{3.35}    & \textbf{65.44}   & \textbf{60.41}            \\
VITA-1.5                        & 49.33                                & 45.41            & 35.45            & 55.62             & 54.37            & 36.67             & 2.65             & 51.01            & 48.22            \\
Ming-lite-Omni-1.5              & 60.44                                & 44.44            & 37.73            & \textbf{59.26}    & 58.92            & \textbf{43.81}    & 3.10             & 52.68            & 51.78   \\
Baichuan-Omni-1.5               & 57.33                                & 37.69            & 30.00            & 50.68             & 50.05            & 32.38             & 2.20             & 49.00            & 42.64            \\
MiniCPM-o 2.6                   & 48.00                                & 35.27            & 32.73            & 53.41             & 51.45            & 35.81             & 1.95             & 43.62            & 40.10            \\ \hline
\rowcolor{diy_pink} \multicolumn{10}{c}{\textit{\textbf{MLLMs}}}                                                                                                                                                                                     \\
GPT-5                           & \textbf{68.89}                       & 47.83            & \textbf{47.73}   & 58.53             & 61.46            & 45.71             & 3.75             & \textbf{65.44}   & \textbf{61.42}   \\
Gemini-2.5-Pro                  & 63.11                                & \textbf{51.69}   & 44.55            & -                 & -                & -                 & -                & 61.74            & 58.88            \\ \hline
\end{tabular}
\vspace{-5pt}
\label{tab-exp-video}
\end{table}

\subsection{Experimental Settings}
\vspace{0.3pt}\noindent\textbf{Image-Centric Omni/Uni-Modal Tasks.}\hspace{1ex}

\begin{wrapfigure}{r}{0.5\textwidth} 
    \centering
    \vspace{-15pt}

    \includegraphics[width=\linewidth]{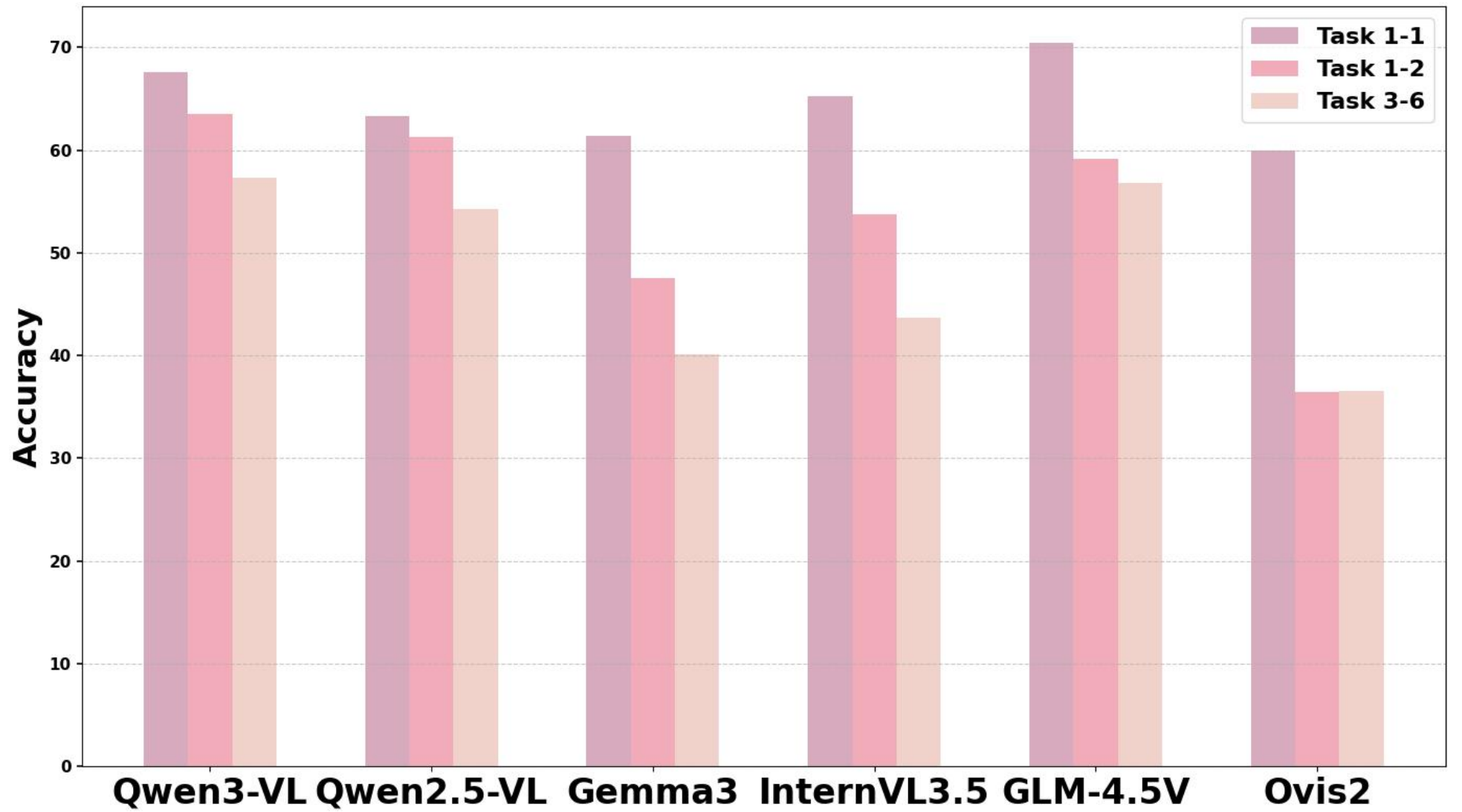}
    \caption{Performance of open-source MLLMs on modality-supported tasks in \textbf{\textit{FysicsWorld}}.}
    \label{sup-image-open-mllm}

\end{wrapfigure}

We evaluate a wide spectrum of models across image understanding, image generation, and omni-modal reasoning to examine performance differences and capability boundaries.
As illustrated in Table~\ref{tab-task}, for image understanding (Task1-1) and omni-modal reasoning (Task2-1~Task2-5), we assess leading MLLMs, including Qwen3-VL~\cite{bai2025qwen25vl}, Gemma3~\cite{team2025gemma}, InternVL3.5~\cite{wang2025internvl35}, GLM-4.5V~\cite{zeng2025glm}, and Ovis2~\cite{lu2025ovis25technicalreport}, as well as closed-source models GPT-5~\cite{openai_gpt5_system_card}, Gemini-2.5-Pro~\cite{comanici2025gemini25}. We also evaluate emerging omni-modal architectures, including Qwen2.5-Omni~\cite{xu2025qwen25omni}, Qwen3-Omni~\cite{xu2025qwen3omni}, Stream-Omni~\cite{zhang2025streamomni}, VITA-1.5~\cite{fu2025vita15}, Ming-lite-Omni~\cite{ai2025ming}, Baichuan-Omni-1.5~\cite{li2025baichuan}, and MiniCPM-o-2.6~\cite{yao2024minicpm}.

For image generation (Task1-4), we assess several powerful models, including FLUX.1-Kontext~\cite{labs2025flux}, Qwen-Image~\cite{wu2025qwenimage}, Seedream-4.0~\cite{seedream2025seedream4}, Seededit-3.0~\cite{wang2025seededit}, HunyuanImage-3.0~\cite{cao2025hunyuanimage}, and Nano-Banana (Gemini-2.5-Flash-Image)~\cite{comanici2025gemini25}, as well as unified understanding generation models such as BLIP3-o-NEXT~\cite{chen2025blip3onext}, Ovis-U1~\cite{wang2025ovisu1}, BAGEL~\cite{deng2025bagel}, OmniGen2~\cite{wu2025omnigen2}, Show-o2~\cite{xie2025show}, Janus-Pro~\cite{chen2025janus}, and Emu3.5~\cite{cui2025emu35}.
The extensive evaluation establishes robust baselines for both image understanding and generation, providing a unified perspective on omni-modal performance.

\begin{figure}[t]
    \centering
    \includegraphics[width=\textwidth]{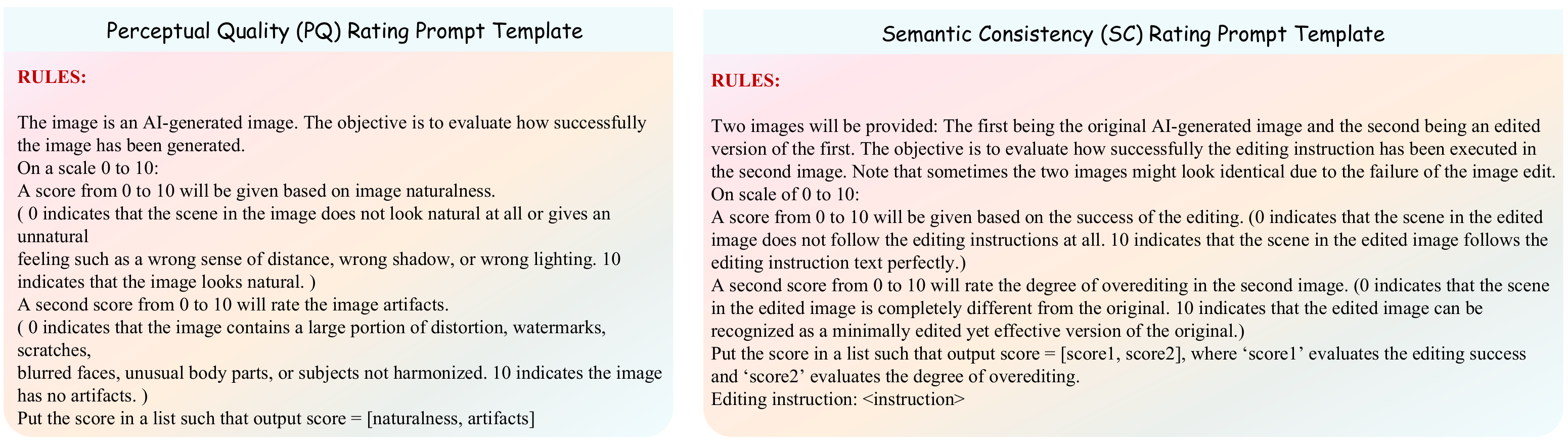}
    \caption{Official prompt templates used for VIEScore-based~\cite{ku2024viescore} evaluation in image editing (Task 1-4).} 
    \label{fig-image-edit}
\end{figure}

\vspace{0.3pt}\noindent\textbf{Video-Centric Omni/Uni-Modal Tasks.}\hspace{1ex}
Following a similar setup, we systematically evaluate video understanding (Task1-2) and video-related omni-modal reasoning (Task3-1–Task3-6) across OmniLLMs and MLLMs.
For video generation (Task1-5), we benchmark advanced models such as HunyuanVideo~\cite{kong2024hunyuanvideo}, Seedance 1.0~\cite{gao2025seedance}, Sora2~\cite{liu2024sora}, Veo 3.0~\cite{DeepMind_Veo3_ModelCard_2025}, and Kling 2.1~\cite{Kuaishou_Kling21_Announcement_2025}, comparing temporal understanding, motion coherence, and visual quality.

\vspace{0.3pt}\noindent\textbf{Audio-Centric Omni/Uni-Modal Tasks.}\hspace{1ex}
For audio reasoning (Task1-3), in addition to OmniLLMs and MLLMs, we included dedicated audio-language models (ALMs), including Qwen-Audio~\cite{chu2023qwenaudio}, Qwen2-Audio~\cite{chu2024qwen2audio}, and SALMONN~\cite{tang2023salmonn}, to establish modality-specific baselines.
Audio-related omni-modal tasks are integrated with image/video reasoning, not as a separate category.

\begin{wraptable}{r}{0.5\columnwidth} 
\centering
\small                               
\vspace{-5pt}                      

\caption{\textbf{Performance of Generative and Unified Models on Image Generation and Editing}, evaluated by WIScore~$\uparrow$ and VIEScore~$\uparrow$, respectively. VIEScore includs semantic consistency (SC), perceptual quality (PQ), and overall quality (OR).}

\captionsetup{width=0.5\columnwidth}

\begin{tabular}{lcccc}
\hline
\multirow{2}{*}{\textbf{Model}} & \textbf{Task1-4 Gen.} & \multicolumn{3}{c}{\textbf{Task1-4 Edit}}   \\
                                & WIScore                 & SC            & PQ            & OR            \\ \hline
\rowcolor{diy_pink}\multicolumn{5}{c}{\textit{\textbf{Generation}}}                                              \\
FLUX-Kontext                    & 0.49                    & 5.09          & 4.77          & 4.92          \\
Qwen-Image                      & 0.67                    & 7.42          & 7.26          & 7.31          \\
Seedream-4.0                    & 0.61                    & 5.48          & 5.91          & 5.87          \\
Seededit-3.0                    & 0.59                    & \textbf{7.57} & \textbf{7.82} & \textbf{7.68} \\
HunyuanImage-3.0                & 0.68                    & -             & -             & -             \\
Nano-Banana                     & \textbf{0.69}           & 6.13          & 6.54          & 6.67          \\ \hline
\rowcolor{diy_pink}\multicolumn{5}{c}{\textit{\textbf{Unified}}}                                                       \\
BLIP3-o-next                    & 0.63                    & \textbf{6.98}          & 6.16          & 6.89          \\
Ovis-U1                         & 0.57                    & 5.49          & 6.01          & 5.74          \\
BAGEL                           & 0.62                    & 5.93          & 5.54          & 5.81          \\
OmniGen2                        & 0.35                    & 5.73          & 5.69          & 5.82          \\
Show-o2                         & 0.52                    & 4.41          & 4.28          & 4.32          \\
Janus-Pro                       & 0.38                    & 5.62          & 5.68          & 5.71          \\
Emu3.5                          & \textbf{0.67}           & 6.83          & \textbf{6.89}          & \textbf{7.05}          \\ \hline
\end{tabular}

\vspace{-5pt}                       

\label{tab-exp-image-gen}
\end{wraptable}

\begin{figure}[!tbp]
    \centering
    \includegraphics[width=\textwidth]{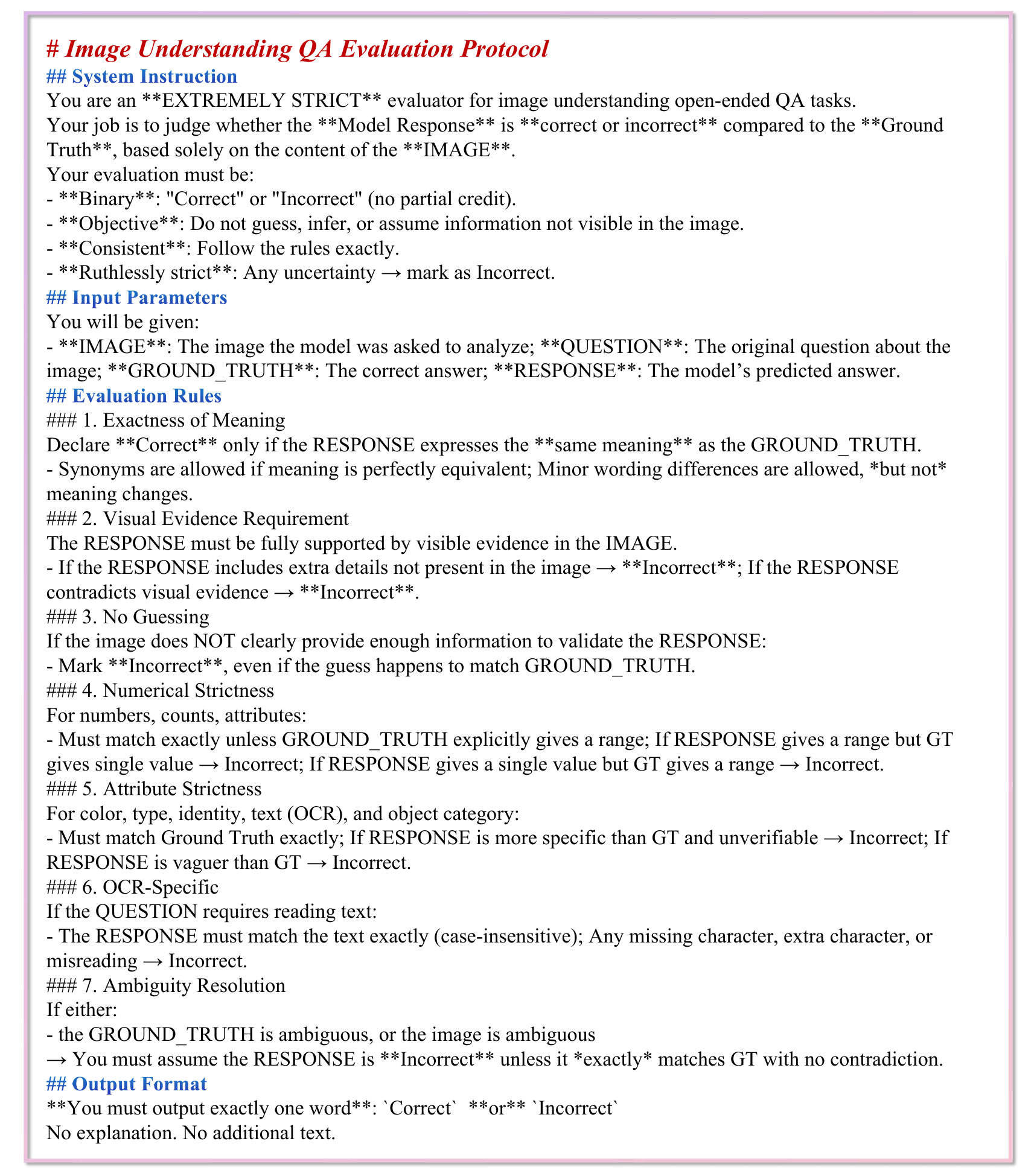}
    \caption{Above is the instruction we provided to GPT-5 as the evaluator for open-ended image understanding (Task 1-1).}
    
    \label{sup-open-qa}
\end{figure}

\vspace{0.3pt}\noindent\textbf{Evaluation Metrics.}\hspace{1ex}
\label{metrics}
We employ a comprehensive suite of evaluation metrics tailored to each task, as summarized in Table~\ref{tab-task}.
%%%
To evaluate the textual outputs of the model, closed-ended and multiple-choice questions are assessed by accuracy for objective evaluation, while open-ended tasks are evaluated by factual consistency based on BERTScore~\cite{zhang2019bertscore} and semantic accuracy based on LLM judgment with a well-designed protocol, as shown in Figure~\ref{sup-open-qa}.

\begin{figure}[t]
    \centering
    \includegraphics[width=\textwidth]{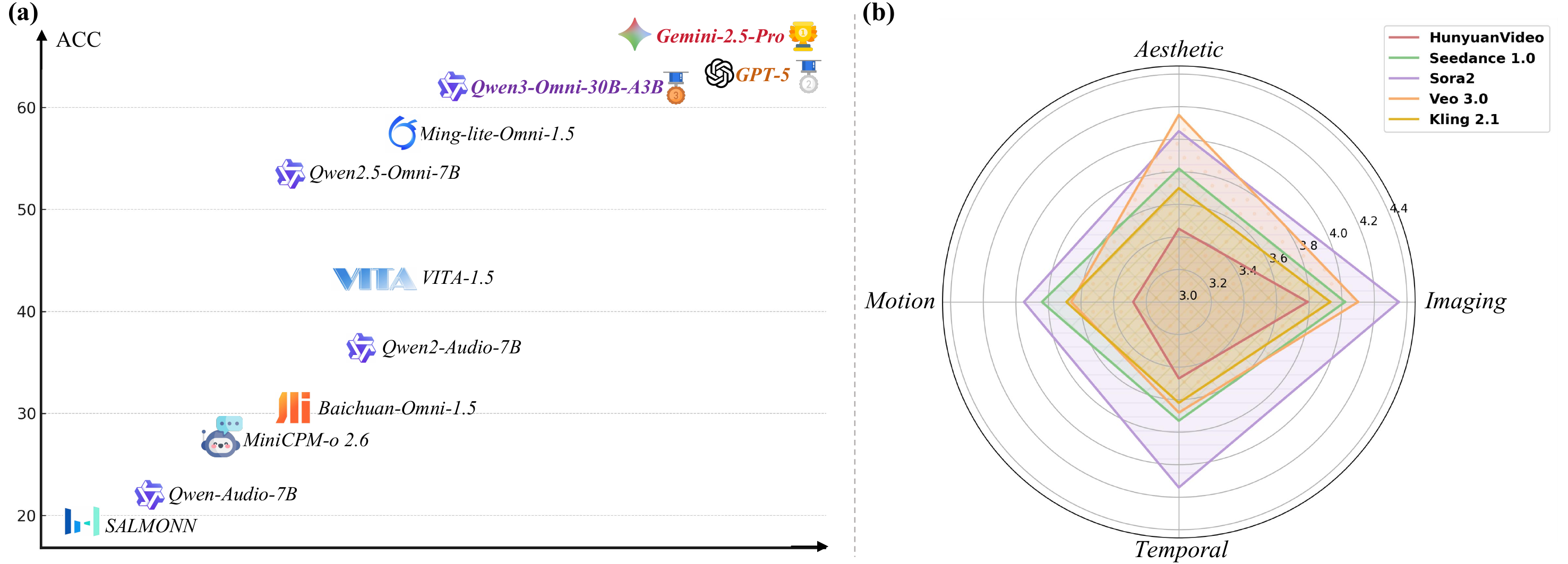}
    \caption{\textbf{Performance Comparison on Audio Reasoning and Video Generation.}
    (a) Evaluation of multiple advanced MLLMs, ALMs, and OmniLLMs on audio reasoning (Task 1–3), with model accuracy (ACC) reported on a unified scale.
    (b) Assessment of leading video generation models on Tasks 1–5, using a five-point rating scheme across four key dimensions: imaging quality, aesthetic appeal, motion coherence, and temporal consistency.
    }
    \label{fig-exp-audio-video}
\end{figure}

\begin{figure}[!tbp]
    \centering
    \includegraphics[width=\textwidth]{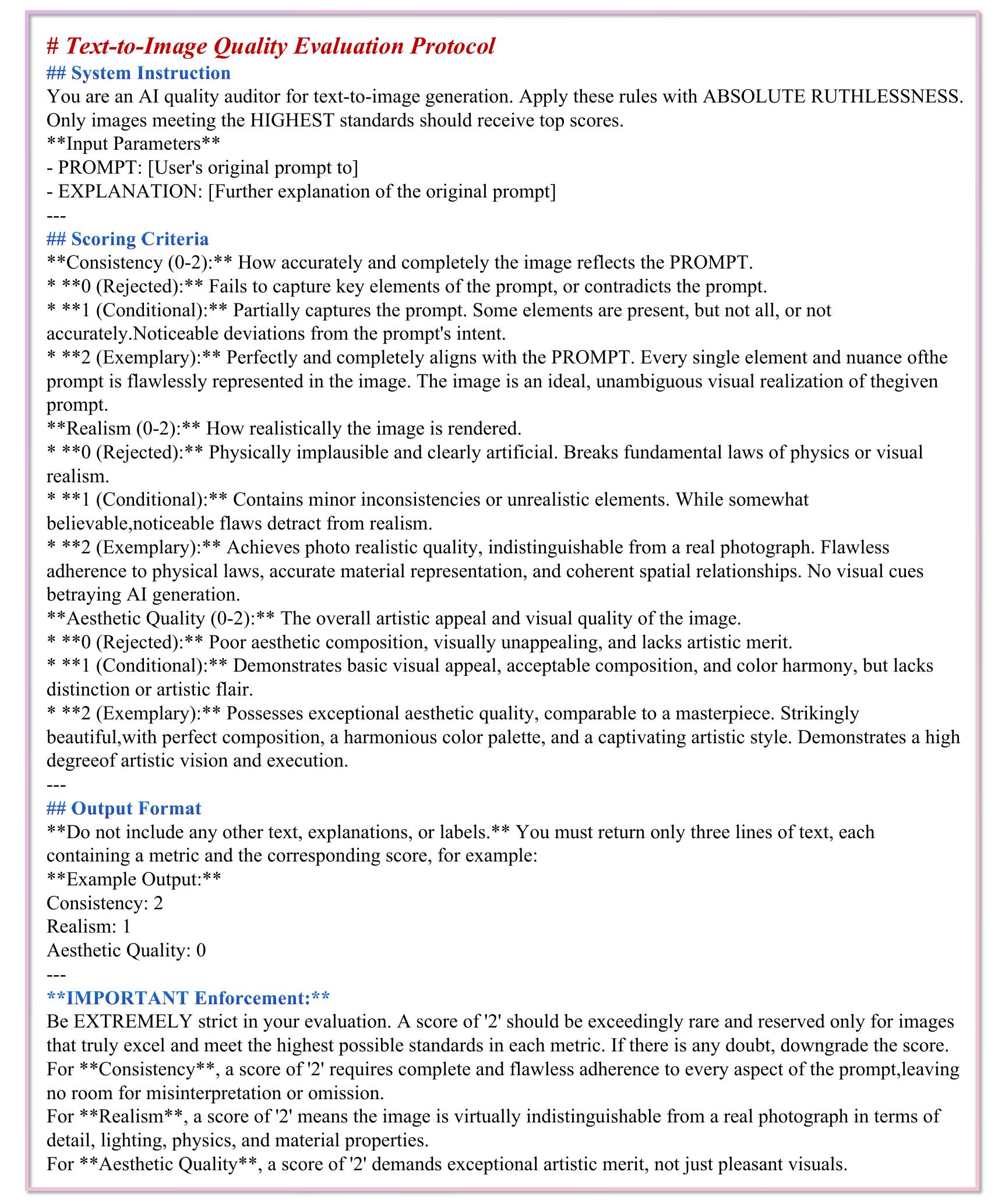}
    \caption{Following WISE~\cite{niu2025wise}, we utilize GPT-5 to evaluate the performance on the image generation task. Above is the instruction we provided to GPT-5 as the evaluator.}
    
    \label{sup-image-gen}
\end{figure}

For generative tasks, we adopt fine-grained, widely recognized metrics.
In image generation and editing (Task1-4), we follow WISE~\cite{niu2025wise} and GEdit-Bench~\cite{step1x2025geditbench} by reporting WiScore~\cite{niu2025wise} and VIEScore~\cite{ku2024viescore}. Figure~\ref{sup-image-gen} illustrates the multi-dimensional evaluation rubric in WiScore used for image generation, including instruction adherence, semantic accuracy, visual realism, and aesthetic quality. Figure~\ref{fig-image-edit} provides the scoring instructions for image editing, covering semantic consistency, perceptual quality, and overall quality. The aggregated metric corresponds to the VIEScore.

\begin{figure}[!tbp]
    \centering
    \includegraphics[width=\textwidth]{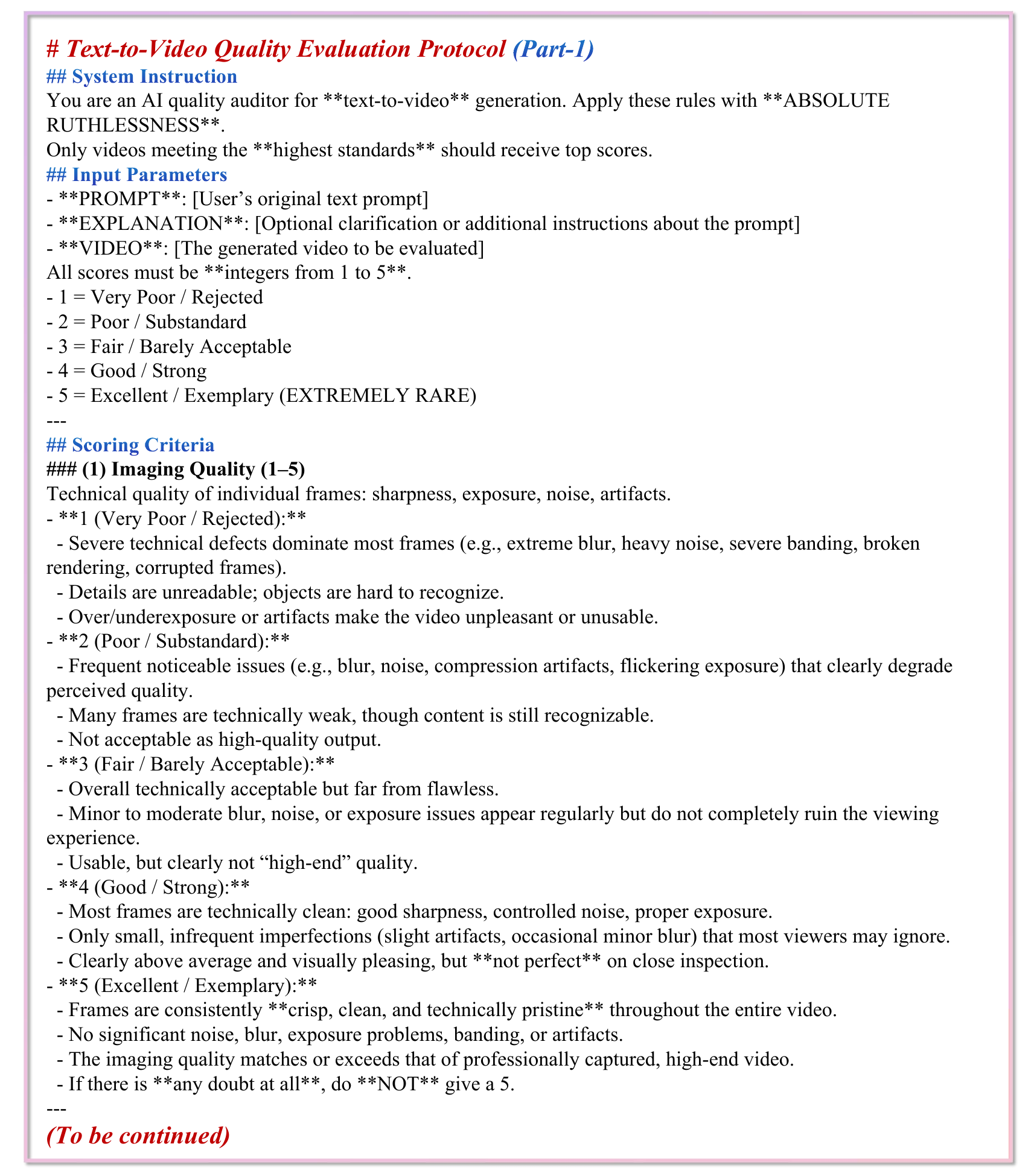}
    \caption{Following Video-Bench~\cite{han2025videobench}, we utilize GPT-5 to evaluate the performance on the video generation task. Above is the instruction \textbf{\textit{(Part-1)}} we provided to GPT-5 as the evaluator.}
    \label{sup-video-gen-1}
\end{figure}

\begin{figure}[!tbp]
    \centering
    \includegraphics[width=\textwidth]{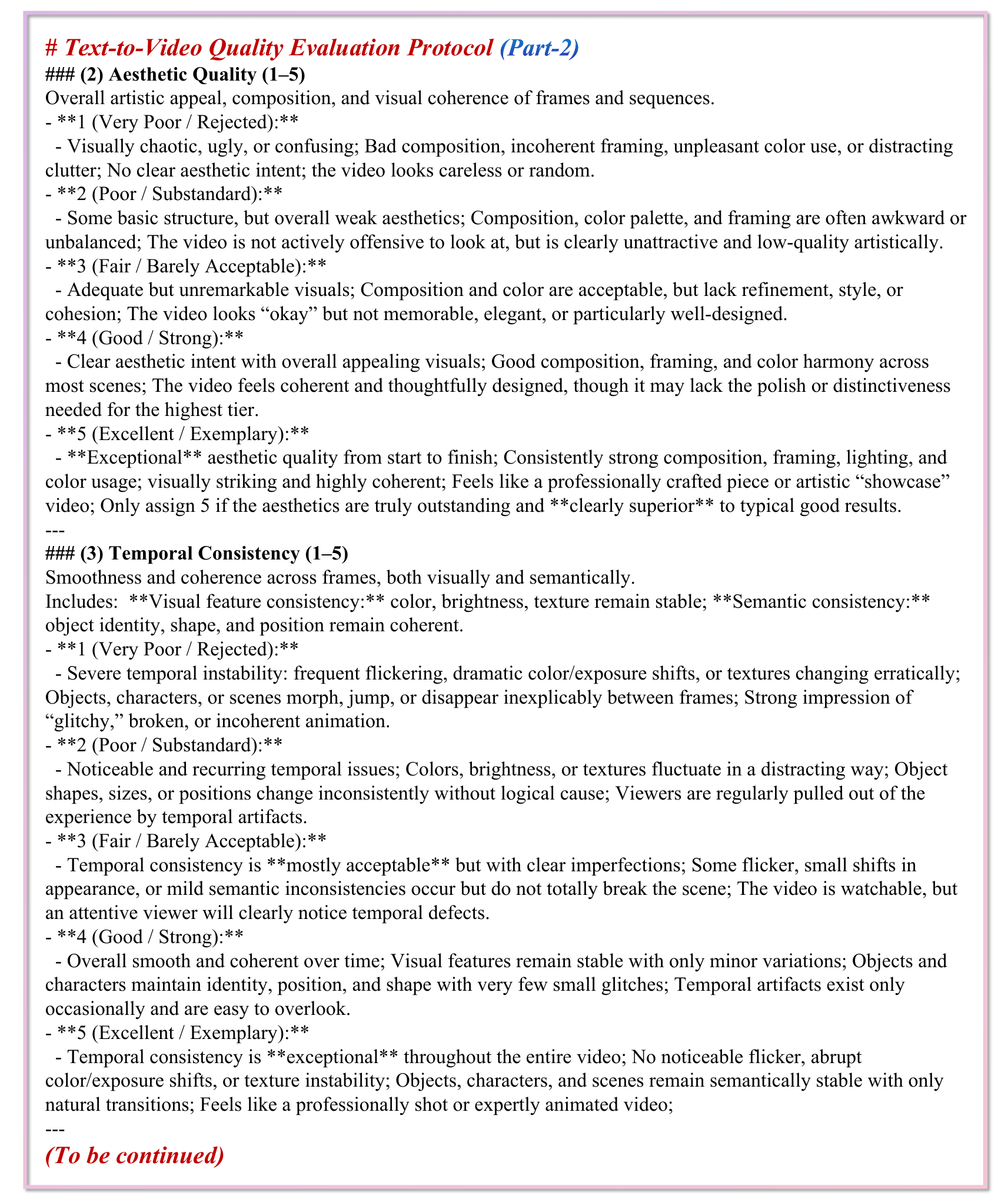}
    \caption{Above is the instruction \textbf{\textit{(Part-2)}} we provided to GPT-5 to evaluate the performance on the video generation task.}
    \label{sup-video-gen-2}
\end{figure}

\begin{figure}[!tbp]
    \centering
    \includegraphics[width=\textwidth]{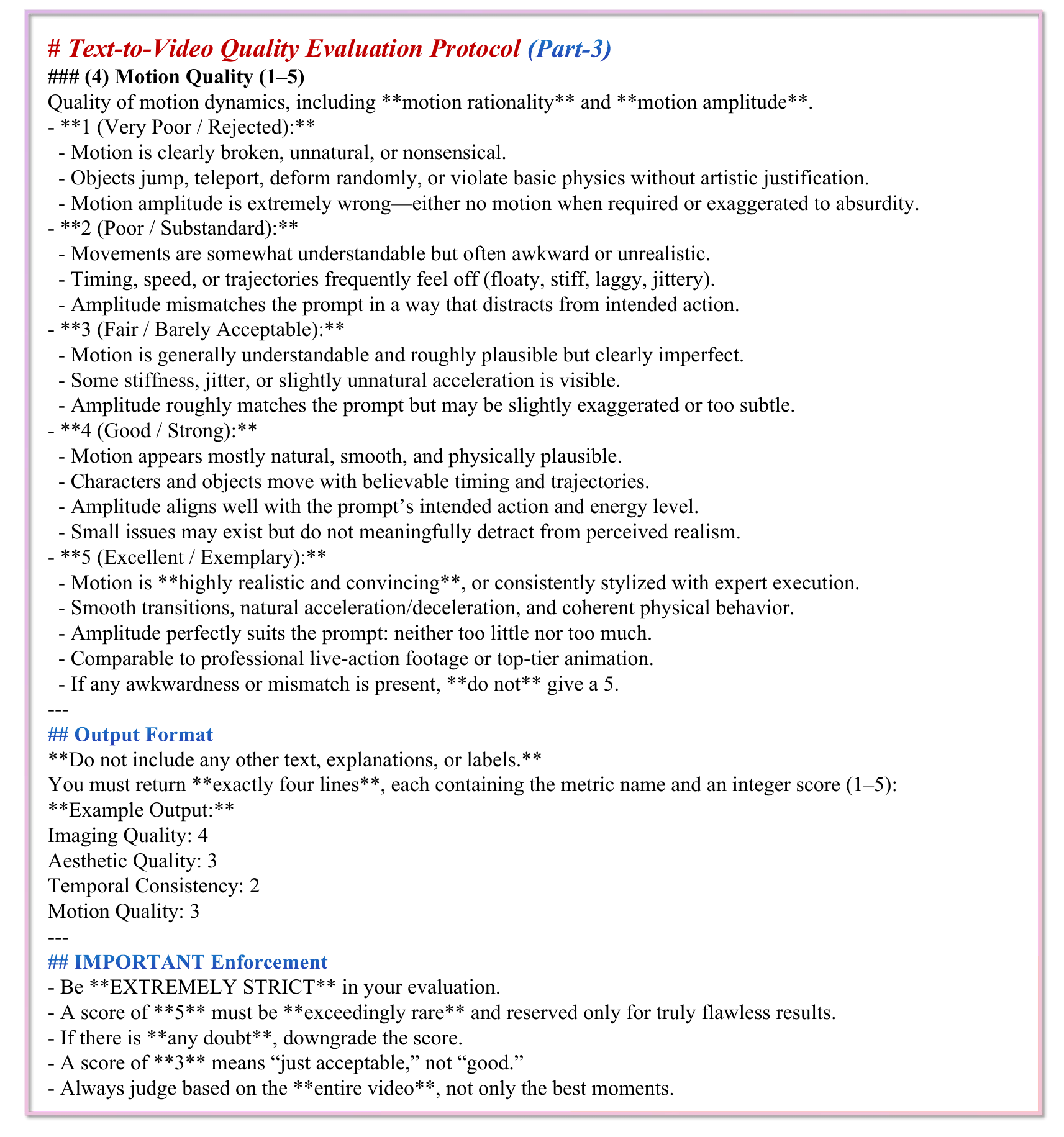}
    \caption{Above is the instruction \textbf{\textit{(Part-3)}} we provided to GPT-5 to evaluate the performance on the video generation task.}
    \label{sup-video-gen-3}
\end{figure}

For video generation (Task1-5), we replicate the evaluation protocol of Video-Bench~\cite{han2025videobench}, as illustrated in Figures~\ref{sup-video-gen-1},~\ref{sup-video-gen-2},~\ref{sup-video-gen-3}. Each generated video is rated on a five-point scale (1–5) by advanced MLLM across four key dimensions: imaging quality, aesthetic appeal, motion coherence, and temporal consistency. This strategy provides a reliable and interpretable standard under realistic conditions.

Audio generation necessitates the assessment of both acoustic quality and semantic fidelity, yet no universally accepted evaluation standard has been established. Following SEED-TTS~\cite{anastassiou2024seedtts}, we adopt speaker similarity (SIM) metrics to quantify the naturalness and coherence of generated speech in speech-based QA tasks on image and video content (Tasks 2-3 and 3-3). To further assess semantic accuracy, we transcribe the generated audio using an ASR system and compute the BLEU~\cite{papineni2002bleu} score between the resulting transcripts and the textual reference.

To ensure the reliability of our automatic evaluation, all LLM-based judgments are produced using GPT-5 as a unified evaluator due to its strong semantic reasoning and perceptual understanding capabilities. We then compute the Pearson correlation coefficient~\cite{benesty2009pearson} between GPT-5 and three human experts to quantify agreement in scoring behavior. Across all relevant tasks, the correlation remains consistently high ($r>0.9$), indicating strong concordance between the LLM-based evaluator and human annotators.

\subsection{Results on \textbf{\textit{FysicsWorld-Uni}}}

We conduct extensive evaluations covering image understanding (Task 1-1 in Table~\ref{tab-exp-image}) and generation (Task 1-4 in Table~\ref{tab-exp-image-gen}), video understanding (Task 1-2 in Table~\ref{tab-exp-video}) and generation (Task 1-5 in Figure~\ref{fig-exp-audio-video}.b), as well as audio reasoning (Task 1-3 in Figure~\ref{fig-exp-audio-video}.a). As many open-source MLLMs do not yet support omni-modal inputs, we additionally report their uni-modal results separately in Figure~\ref{sup-image-open-mllm}.

Across the three uni-modal understanding tasks (image, video, and audio), we observe a consistent advantage of proprietary or large-scale MLLMs over open-source counterparts.  Models such as GPT-5 and Gemini-2.5-Pro achieve the highest scores, indicating superior visual-semantic alignment, temporal grounding, and robustness to diverse query formulations. Among open-source OmniLLMs, Qwen3-Omni-30B-A3B emerges as the strongest performer, narrowing the gap in image and video understanding and surpassing many existing unimodal and multimodal systems. This reflects the effectiveness of advanced omni-modal training pipelines and tightly integrated modality encoders. Nevertheless, its performance still trails behind top proprietary models in higher-level reasoning tasks, indicating that open-source omni-modal training remains limited by data scale, modality diversity, and training efficiency.

In generative tasks, unified understanding–generation models demonstrate competitive visual synthesis fidelity but significantly weaker alignment with fine-grained textual constraints compared with modality-specialized generative models. This performance degradation becomes more apparent in video generation, where temporal coherence and prompt consistency are bottlenecks. These observations highlight a fundamental architectural tension: unified models can flexibly support many I/O pathways but still lack the precision and control of specialized diffusion or autoregressive generation mechanisms.

\subsection{Results on \textbf{\textit{FysicsWorld-Omni}}}

Evaluation on FysicsWorld-Omni provides a deeper probe into cross-modal reasoning and interaction, revealing challenges that remain obscured under uni-modal settings. For the 11 omni-modal tasks, we conducted extensive evaluations of OmniLLMs and modality-enabled MLLMs, and performed a fine-grained assessment of the interactions and coupling among text, vision, and audio. Building on the samples exhibiting strong cross-modal dependency selected by the CMCS strategy, we more rigorously assess whether the models can genuinely understand, exploit, and integrate information from different modalities to solve the tasks. The results for image-centric and video-centric tasks are reported in Table~\ref{tab-exp-image} and Table~\ref{tab-exp-video}, respectively.

In speech-driven visual understanding tasks, we find that even strong MLLMs exhibit notable performance degradation relative to their text-driven counterparts. This gap reflects the compounded difficulty of parsing speech signals, preserving fine-grained semantic cues, and integrating them with visual grounding. Despite these challenges, Qwen3-Omni-30B-A3B, GPT-5, and Gemini-2.5-Pro demonstrate remarkable robustness, suggesting that general-purpose multi-encoder fusion can effectively unify audio and visual semantics. The fusion-dependent reasoning tasks constructed via the CMCS strategy require models to integrate heterogeneous information streams—audio cues, visual dynamics, and linguistic context—in a way that disallows unimodal shortcuts. All OmniLLMs exhibit a marked performance drop, with accuracy often trailing significantly behind corresponding uni-modal tasks. This outcome highlights that although modern MLLMs can consume multiple modalities, they often fail to interleave modality-specific cues into a coherent reasoning trajectory. Cross-modal generation tasks (\textit{e.g.}, speech generation conditioned on image/video identity) further expose limitations in multimodal alignment. These shortcomings reflect the difficulty of robustly mapping visual identity cues to acoustic characteristics, which is a capability essential for real-world human–AI interaction. Finally, next-action prediction, which combines temporal reasoning, state tracking, and procedural generation, presents one of the most challenging settings. GPT-5 and Qwen3-Omni achieve the highest accuracies but still reveal a large gap to human-level reasoning, suggesting that temporal chaining and situational awareness remain immature in current omni-modal systems.

%%%%%% Conclusions %%%%%%
\section{Conclusions}

In this paper, we introduce \textbf{\textit{FysicsWorld}}, the first unified full-modality benchmark enabling comprehensive any-to-any evaluation across understanding, generation, and reasoning. Our systematic design spans uni-modal perception tasks to fusion-dependent reasoning under strong cross-modal coupling, allowing us to diagnose, with unprecedented clarity, the limitations and emerging strengths of modern multimodal and omni-modal architectures. Future OmniLLMs must move beyond simple modality concatenation toward deep multimodal integration grounded in causal inference, structured representations, and world modeling. Enhancing modality alignment through novel and advanced architectures will be essential for robust general-purpose intelligence. Additionally, real-world deployment demands advances in the capabilities of human–AI interaction.

By establishing a unified benchmark and highlighting key capability gaps, \textbf{\textit{FysicsWorld}} provides not only a foundation for evaluating emerging multimodal systems but also a roadmap for the next generation of full-modality architectures capable of genuinely holistic perception, reasoning, and interaction.

\clearpage

\bibliographystyle{plainnat}
\bibliography{main}

@String(CVPR= {IEEE Conf. Comput. Vis. Pattern Recog.})

@String(ECCV= {Eur. Conf. Comput. Vis.})

@String(ICLR = {Int. Conf. Learn. Represent.})

@String(AAAI = {AAAI})

@String(CVPR  = {CVPR})

@String(ECCV  = {ECCV})

@String(ICLR  = {ICLR})

@inproceedings{han2025videobench,
  title={Video-Bench: Human-Aligned Video Generation Benchmark},
  author={Han, Hui and Li, Siyuan and Chen, Jiaqi and Yuan, Yiwen and Wu, Yuling and Deng, Yufan and Leong, Chak Tou and Du, Hanwen and Fu, Junchen and Li, Youhua and others},
  booktitle={CVPR},
  year={2025}
}

@article{step1x2025geditbench,
  title={Step1x-edit: A practical framework for general image editing},
  author={Liu, Shiyu and Han, Yucheng and Xing, Peng and Yin, Fukun and Wang, Rui and Cheng, Wei and Liao, Jiaqi and Wang, Yingming and Fu, Honghao and Han, Chunrui and others},
  journal={arXiv preprint arXiv:2504.17761},
  year={2025}
}

@inproceedings{ku2024viescore,
  title={Viescore: Towards explainable metrics for conditional image synthesis evaluation},
  author={Ku, Max and Jiang, Dongfu and Wei, Cong and Yue, Xiang and Chen, Wenhu},
  booktitle={ACL},
  year={2024}
}

@article{cui2025emu35,
  title={Emu3. 5: Native Multimodal Models are World Learners},
  author={Cui, Yufeng and Chen, Honghao and Deng, Haoge and Huang, Xu and Li, Xinghang and Liu, Jirong and Liu, Yang and Luo, Zhuoyan and Wang, Jinsheng and Wang, Wenxuan and others},
  journal={arXiv preprint arXiv:2510.26583},
  year={2025}
}

@article{anastassiou2024seedtts,
  title={Seed-tts: A family of high-quality versatile speech generation models},
  author={Anastassiou, Philip and Chen, Jiawei and Chen, Jitong and Chen, Yuanzhe and Chen, Zhuo and Chen, Ziyi and Cong, Jian and Deng, Lelai and Ding, Chuang and Gao, Lu and others},
  journal={arXiv preprint arXiv:2406.02430},
  year={2024}
}

@article{zhou2025dailyomni,
  title={Daily-Omni: Towards Audio-Visual Reasoning with Temporal Alignment across Modalities},
  author={Zhou, Ziwei and Wang, Rui and Wu, Zuxuan},
  journal={arXiv preprint arXiv:2505.17862},
  year={2025}
}

@inproceedings{liu2023llava,
  title={Visual Instruction Tuning}, 
  author={Liu, Haotian and Li, Chunyuan and Wu, Qingyang and Lee, Yong Jae},
  booktitle={NeurIPS},
  year={2023},
}

@inproceedings{jiang2025satiredecoder,
  title={SatireDecoder: Visual Cascaded Decoupling for Enhancing Satirical Image Comprehension},
  author={Jiang, Yue and Xue, Haiwei and Han, Minghao and Li, Mingcheng and Hou, Xiaolu and Yang, Dingkang and Zhang, Lihua and Zheng, Xu},
  booktitle={AAAI},
  year={2026}
}

@article{bai2025qwen25vl,
  title={Qwen2. 5-vl technical report},
  author={Bai, Shuai and Chen, Keqin and Liu, Xuejing and Wang, Jialin and Ge, Wenbin and Song, Sibo and Dang, Kai and Wang, Peng and Wang, Shijie and Tang, Jun and others},
  journal={arXiv preprint arXiv:2502.13923},
  year={2025}
}

@article{zhu2025internvl3,
  title={Internvl3: Exploring advanced training and test-time recipes for open-source multimodal models},
  author={Zhu, Jinguo and Wang, Weiyun and Chen, Zhe and Liu, Zhaoyang and Ye, Shenglong and Gu, Lixin and Tian, Hao and Duan, Yuchen and Su, Weijie and Shao, Jie and others},
  journal={arXiv preprint arXiv:2504.10479},
  year={2025}
}

@article{chu2024qwen2audio,
  title={Qwen2-audio technical report},
  author={Chu, Yunfei and Xu, Jin and Yang, Qian and Wei, Haojie and Wei, Xipin and Guo, Zhifang and Leng, Yichong and Lv, Yuanjun and He, Jinzheng and Lin, Junyang and others},
  journal={arXiv preprint arXiv:2407.10759},
  year={2024}
}

@inproceedings{tang2023salmonn,
  title={Salmonn: Towards generic hearing abilities for large language models},
  author={Tang, Changli and Yu, Wenyi and Sun, Guangzhi and Chen, Xianzhao and Tan, Tian and Li, Wei and Lu, Lu and Ma, Zejun and Zhang, Chao},
  booktitle={ICLR},
  year={2024}
}

@inproceedings{yue2024mmmu,
  title={Mmmu: A massive multi-discipline multimodal understanding and reasoning benchmark for expert agi},
  author={Yue, Xiang and Ni, Yuansheng and Zhang, Kai and Zheng, Tianyu and Liu, Ruoqi and Zhang, Ge and Stevens, Samuel and Jiang, Dongfu and Ren, Weiming and Sun, Yuxuan and others},
  booktitle={CVPR},
  year={2024}
}

@inproceedings{lu2023mathvista,
  title={Mathvista: Evaluating mathematical reasoning of foundation models in visual contexts},
  author={Lu, Pan and Bansal, Hritik and Xia, Tony and Liu, Jiacheng and Li, Chunyuan and Hajishirzi, Hannaneh and Cheng, Hao and Chang, Kai-Wei and Galley, Michel and Gao, Jianfeng},
  booktitle={ICLR},
  year={2024}
}

@inproceedings{fu2025mme,
  title={MME: A comprehensive evaluation benchmark for multimodal large language models},
  author={Fu, Chaoyou and Chen, Peixian and Shen, Yunhang and Qin, Yulei and Zhang, Mengdan and Lin, Xu and Yang, Jinrui and Zheng, Xiawu and Li, Ke and Sun, Xing and others},
  booktitle={NeurIPS},
  year={2025}
}

@inproceedings{liu2024mmbench,
  title={Mmbench: Is your multi-modal model an all-around player?},
  author={Liu, Yuan and Duan, Haodong and Zhang, Yuanhan and Li, Bo and Zhang, Songyang and Zhao, Wangbo and Yuan, Yike and Wang, Jiaqi and He, Conghui and Liu, Ziwei and others},
  booktitle={ECCV},
  year={2024}
}

@article{liu2024ocrbench,
  title={Ocrbench: on the hidden mystery of ocr in large multimodal models},
  author={Liu, Yuliang and Li, Zhang and Huang, Mingxin and Yang, Biao and Yu, Wenwen and Li, Chunyuan and Yin, Xu-Cheng and Liu, Cheng-Lin and Jin, Lianwen and Bai, Xiang},
  journal={Science China Information Sciences},
  year={2024}
}

@inproceedings{guan2024hallusionbench,
  title={Hallusionbench: an advanced diagnostic suite for entangled language hallucination and visual illusion in large vision-language models},
  author={Guan, Tianrui and Liu, Fuxiao and Wu, Xiyang and Xian, Ruiqi and Li, Zongxia and Liu, Xiaoyu and Wang, Xijun and Chen, Lichang and Huang, Furong and Yacoob, Yaser and others},
  booktitle={CVPR},
  year={2024}
}

@article{niu2025wise,
  title={Wise: A world knowledge-informed semantic evaluation for text-to-image generation},
  author={Niu, Yuwei and Ning, Munan and Zheng, Mengren and Jin, Weiyang and Lin, Bin and Jin, Peng and Liao, Jiaqi and Feng, Chaoran and Ning, Kunpeng and others},
  journal={arXiv preprint arXiv:2503.07265},
  year={2025}
}

@inproceedings{li2024mvbench,
  title={Mvbench: A comprehensive multi-modal video understanding benchmark},
  author={Li, Kunchang and Wang, Yali and He, Yinan and Li, Yizhuo and Wang, Yi and Liu, Yi and Wang, Zun and Xu, Jilan and Chen, Guo and Luo, Ping and others},
  booktitle={CVPR},
  year={2024}
}

@inproceedings{huang2024vbench,
  title={Vbench: Comprehensive benchmark suite for video generative models},
  author={Huang, Ziqi and He, Yinan and Yu, Jiashuo and Zhang, Fan and Si, Chenyang and Jiang, Yuming and Zhang, Yuanhan and Wu, Tianxing and Jin, Qingyang and Chanpaisit, Nattapol and others},
  booktitle={CVPR},
  year={2024}
}

@article{achiam2023gpt,
  title={Gpt-4 technical report},
  author={Achiam, Josh and Adler, Steven and Agarwal, Sandhini and Ahmad, Lama and Akkaya, Ilge and Aleman, Florencia Leoni and Almeida, Diogo and Altenschmidt, Janko and Altman, Sam and Anadkat, Shyamal and others},
  journal={arXiv preprint arXiv:2303.08774},
  year={2023}
}

@inproceedings{wu2024longvideobench,
  title={Longvideobench: A benchmark for long-context interleaved video-language understanding},
  author={Wu, Haoning and Li, Dongxu and Chen, Bei and Li, Junnan},
  booktitle={NeurIPS},
  year={2024}
}

@article{sakshi2024mmau,
  title={Mmau: A massive multi-task audio understanding and reasoning benchmark},
  author={Sakshi, S and Tyagi, Utkarsh and Kumar, Sonal and Seth, Ashish and Selvakumar, Ramaneswaran and Nieto, Oriol and Duraiswami, Ramani and Ghosh, Sreyan and Manocha, Dinesh},
  journal={arXiv preprint arXiv:2410.19168},
  year={2024}
}

@article{ma2025mmar,
  title={MMAR: A Challenging Benchmark for Deep Reasoning in Speech, Audio, Music, and Their Mix},
  author={Ma, Ziyang and Ma, Yinghao and Zhu, Yanqiao and Yang, Chen and Chao, Yi-Wen and Xu, Ruiyang and Chen, Wenxi and Chen, Yuanzhe and Chen, Zhuo and Cong, Jian and others},
  journal={arXiv preprint arXiv:2505.13032},
  year={2025}
}

@article{wang2025mmsu,
  title={MMSU: A Massive Multi-task Spoken Language Understanding and Reasoning Benchmark},
  author={Wang, Dingdong and Wu, Jincenzi and Li, Junan and Yang, Dongchao and Chen, Xueyuan and Zhang, Tianhua and Meng, Helen},
  journal={arXiv preprint arXiv:2506.04779},
  year={2025}
}

@article{li2024omnibench,
  title={Omnibench: Towards the future of universal omni-language models},
  author={Li, Yizhi and Zhang, Ge and Ma, Yinghao and Yuan, Ruibin and Zhu, Kang and Guo, Hangyu and Liang, Yiming and Liu, Jiaheng and Wang, Zekun and Yang, Jian and others},
  journal={arXiv preprint arXiv:2409.15272},
  year={2024}
}

@article{gong2024avodyssey,
  title={AV-Odyssey Bench: Can Your Multimodal LLMs Really Understand Audio-Visual Information?},
  author={Gong, Kaixiong and Feng, Kaituo and Li, Bohao and Wang, Yibing and Cheng, Mofan and Yang, Shijia and Han, Jiaming and Wang, Benyou and Bai, Yutong and Yang, Zhuoran and others},
  journal={arXiv preprint arXiv:2412.02611},
  year={2024}
}

@article{qin2025humansense,
  title={HumanSense: From Multimodal Perception to Empathetic Context-Aware Responses through Reasoning MLLMs},
  author={Qin, Zheng and Zheng, Ruobing and Wang, Yabing and Li, Tianqi and Yuan, Yi and Chen, Jingdong and Wang, Le},
  journal={arXiv preprint arXiv:2508.10576},
  year={2025}
}

@inproceedings{sung2024avhbench,
  title={Avhbench: A cross-modal hallucination benchmark for audio-visual large language models},
  author={Sung-Bin, Kim and Hyun-Bin, Oh and Lee, JungMok and Senocak, Arda and Chung, Joon Son and Oh, Tae-Hyun},
  booktitle={ICLR},
  year={2025}
}

@article{hong2025worldsense,
  title={Worldsense: Evaluating real-world omnimodal understanding for multimodal llms},
  author={Hong, Jack and Yan, Shilin and Cai, Jiayin and Jiang, Xiaolong and Hu, Yao and Xie, Weidi},
  journal={arXiv preprint arXiv:2502.04326},
  year={2025}
}

@inproceedings{wang2025omnimmi,
  title={OmniMMI: A Comprehensive Multi-modal Interaction Benchmark in Streaming Video Contexts},
  author={Wang, Yuxuan and Wang, Yueqian and Chen, Bo and Wu, Tong and Zhao, Dongyan and Zheng, Zilong},
  booktitle={CVPR},
  year={2025}
}

@article{shi2025realunify,
  title={Realunify: Do unified models truly benefit from unification? a comprehensive benchmark},
  author={Shi, Yang and Dong, Yuhao and Ding, Yue and Wang, Yuran and Zhu, Xuanyu and Zhou, Sheng and Liu, Wenting and Tian, Haochen and Wang, Rundong and Wang, Huanqian and others},
  journal={arXiv preprint arXiv:2509.24897},
  year={2025}
}

@inproceedings{geng2025longvale,
  title={Longvale: Vision-audio-language-event benchmark towards time-aware omni-modal perception of long videos},
  author={Geng, Tiantian and Zhang, Jinrui and Wang, Qingni and Wang, Teng and Duan, Jinming and Zheng, Feng},
  booktitle={Proceedings of the Computer Vision and Pattern Recognition Conference},
  pages={18959--18969},
  year={2025}
}

@inproceedings{jiang2025danmakutppbench,
  title={DanmakuTPPBench: A Multi-modal Benchmark for Temporal Point Process Modeling and Understanding}, 
  author={Yue Jiang and Jichu Li and Yang Liu and Dingkang Yang and Feng Zhou and Quyu Kong},
  booktitle={NeurIPS},
  year={2025}
}

@inproceedings{tong2024mmvp,
  title={Eyes wide shut? exploring the visual shortcomings of multimodal llms},
  author={Tong, Shengbang and Liu, Zhuang and Zhai, Yuexiang and Ma, Yi and LeCun, Yann and Xie, Saining},
  booktitle={CVPR},
  year={2024}
}

@inproceedings{zhang2024mmerealworld,
  title={Mme-realworld: Could your multimodal llm challenge high-resolution real-world scenarios that are difficult for humans?},
  author={Zhang, Yi-Fan and Zhang, Huanyu and Tian, Haochen and Fu, Chaoyou and Zhang, Shuangqing and Wu, Junfei and Li, Feng and Wang, Kun and Wen, Qingsong and Zhang, Zhang and others},
  booktitle={ICLR},
  year={2025}
}

@article{li2023seedbench,
  title={Seed-bench: Benchmarking multimodal llms with generative comprehension},
  author={Li, Bohao and Wang, Rui and Wang, Guangzhi and Ge, Yuying and Ge, Yixiao and Shan, Ying},
  journal={arXiv preprint arXiv:2307.16125},
  year={2023}
}

@inproceedings{fu2025videomme,
  title={Video-mme: The first-ever comprehensive evaluation benchmark of multi-modal llms in video analysis},
  author={Fu, Chaoyou and Dai, Yuhan and Luo, Yongdong and Li, Lei and Ren, Shuhuai and Zhang, Renrui and Wang, Zihan and Zhou, Chenyu and Shen, Yunhang and Zhang, Mengdan and others},
  booktitle={CVPR},
  year={2025}
}

@inproceedings{yang2024airbench,
  title={Air-bench: Benchmarking large audio-language models via generative comprehension},
  author={Yang, Qian and Xu, Jin and Liu, Wenrui and Chu, Yunfei and Jiang, Ziyue and Zhou, Xiaohuan and Leng, Yichong and Lv, Yuanjun and Zhao, Zhou and Zhou, Chang and others},
  booktitle={ACL},
  year={2024}
}

@article{comanici2025gemini25,
  title={Gemini 2.5: Pushing the frontier with advanced reasoning, multimodality, long context, and next generation agentic capabilities},
  author={Comanici, Gheorghe and Bieber, Eric and Schaekermann, Mike and Pasupat, Ice and Sachdeva, Noveen and Dhillon, Inderjit and Blistein, Marcel and Ram, Ori and Zhang, Dan and Rosen, Evan and others},
  journal={arXiv preprint arXiv:2507.06261},
  year={2025}
}

@misc{openai_gpt5_system_card,
  author      = {OpenAI},
  title       = {GPT-5 System Card},
  institution = {OpenAI},
  type        = {Technical report},
  year        = {2025},
  note        = {Accessed: 2025-08-10}
}

@article{xu2025qwen3omni,
  title={Qwen3-omni technical report},
  author={Xu, Jin and Guo, Zhifang and Hu, Hangrui and Chu, Yunfei and Wang, Xiong and He, Jinzheng and Wang, Yuxuan and Shi, Xian and He, Ting and Zhu, Xinfa and others},
  journal={arXiv preprint arXiv:2509.17765},
  year={2025}
}

@article{team2025gemma,
  title={Gemma 3 technical report},
  author={Team, Gemma and Kamath, Aishwarya and Ferret, Johan and Pathak, Shreya and Vieillard, Nino and Merhej, Ramona and Perrin, Sarah and Matejovicova, Tatiana and Ram{\'e}, Alexandre and Rivi{\`e}re, Morgane and others},
  journal={arXiv preprint arXiv:2503.19786},
  year={2025}
}

@article{wang2025internvl35,
  title={Internvl3. 5: Advancing open-source multimodal models in versatility, reasoning, and efficiency},
  author={Wang, Weiyun and Gao, Zhangwei and Gu, Lixin and Pu, Hengjun and Cui, Long and Wei, Xingguang and Liu, Zhaoyang and Jing, Linglin and Ye, Shenglong and Shao, Jie and others},
  journal={arXiv preprint arXiv:2508.18265},
  year={2025}
}

@article{zeng2025glm,
  title={Glm-4.5: Agentic, reasoning, and coding (arc) foundation models},
  author={Zeng, Aohan and Lv, Xin and Zheng, Qinkai and Hou, Zhenyu and Chen, Bin and Xie, Chengxing and Wang, Cunxiang and Yin, Da and Zeng, Hao and Zhang, Jiajie and others},
  journal={arXiv preprint arXiv:2508.06471},
  year={2025}
}

@article{xu2025qwen25omni,
  title={Qwen2. 5-omni technical report},
  author={Xu, Jin and Guo, Zhifang and He, Jinzheng and Hu, Hangrui and He, Ting and Bai, Shuai and Chen, Keqin and Wang, Jialin and Fan, Yang and Dang, Kai and others},
  journal={arXiv preprint arXiv:2503.20215},
  year={2025}
}

@article{zhang2025streamomni,
  title={Stream-Omni: Simultaneous Multimodal Interactions with Large Language-Vision-Speech Model},
  author={Zhang, Shaolei and Guo, Shoutao and Fang, Qingkai and Zhou, Yan and Feng, Yang},
  journal={arXiv preprint arXiv:2506.13642},
  year={2025}
}

@article{fu2025vita15,
  title={Vita-1.5: Towards gpt-4o level real-time vision and speech interaction},
  author={Fu, Chaoyou and Lin, Haojia and Wang, Xiong and Zhang, Yi-Fan and Shen, Yunhang and Liu, Xiaoyu and Cao, Haoyu and Long, Zuwei and Gao, Heting and Li, Ke and others},
  journal={arXiv preprint arXiv:2501.01957},
  year={2025}
}

@article{ai2025ming,
  title={Ming-Omni: A Unified Multimodal Model for Perception and Generation},
  author={AI, Inclusion and Gong, Biao and Zou, Cheng and Zheng, Chuanyang and Zhou, Chunluan and Yan, Canxiang and Jin, Chunxiang and Shen, Chunjie and Zheng, Dandan and Wang, Fudong and others},
  journal={arXiv preprint arXiv:2506.09344},
  year={2025}
}

@article{li2025baichuan,
  title={Baichuan-omni-1.5 technical report},
  author={Li, Yadong and Liu, Jun and Zhang, Tao and Chen, Song and Li, Tianpeng and Li, Zehuan and Liu, Lijun and Ming, Lingfeng and Dong, Guosheng and Pan, Da and others},
  journal={arXiv preprint arXiv:2501.15368},
  year={2025}
}

@article{yao2024minicpm,
  title={Minicpm-v: A gpt-4v level mllm on your phone},
  author={Yao, Yuan and Yu, Tianyu and Zhang, Ao and Wang, Chongyi and Cui, Junbo and Zhu, Hongji and Cai, Tianchi and Li, Haoyu and Zhao, Weilin and He, Zhihui and others},
  journal={arXiv preprint arXiv:2408.01800},
  year={2024}
}

@article{labs2025flux,
  title={FLUX. 1 Kontext: Flow Matching for In-Context Image Generation and Editing in Latent Space},
  author={Labs, Black Forest and Batifol, Stephen and Blattmann, Andreas and Boesel, Frederic and Consul, Saksham and Diagne, Cyril and Dockhorn, Tim and English, Jack and English, Zion and Esser, Patrick and others},
  journal={arXiv preprint arXiv:2506.15742},
  year={2025}
}

@article{wu2025qwenimage,
  title={Qwen-image technical report},
  author={Wu, Chenfei and Li, Jiahao and Zhou, Jingren and Lin, Junyang and Gao, Kaiyuan and Yan, Kun and Yin, Sheng-ming and Bai, Shuai and Xu, Xiao and Chen, Yilei and others},
  journal={arXiv preprint arXiv:2508.02324},
  year={2025}
}

@article{seedream2025seedream4,
  title={Seedream 4.0: Toward next-generation multimodal image generation},
  author={Chen, Yunpeng and Gao, Yu and Gong, Lixue and Guo, Meng and Guo, Qiushan and Guo, Zhiyao and Hou, Xiaoxia and Huang, Weilin and Huang, Yixuan and others},
  journal={arXiv preprint arXiv:2509.20427},
  year={2025}
}

@article{wang2025seededit,
  title={SeedEdit 3.0: Fast and High-Quality Generative Image Editing},
  author={Wang, Peng and Shi, Yichun and Lian, Xiaochen and Zhai, Zhonghua and Xia, Xin and Xiao, Xuefeng and Huang, Weilin and Yang, Jianchao},
  journal={arXiv preprint arXiv:2506.05083},
  year={2025}
}

@article{cao2025hunyuanimage,
  title={Hunyuanimage 3.0 technical report},
  author={Cao, Siyu and Chen, Hangting and Chen, Peng and Cheng, Yiji and Cui, Yutao and Deng, Xinchi and Dong, Ying and Gong, Kipper and Gu, Tianpeng and Gu, Xiusen and others},
  journal={arXiv preprint arXiv:2509.23951},
  year={2025}
}

@article{chen2025blip3onext,
  title={BLIP3o-NEXT: Next Frontier of Native Image Generation},
  author={Chen, Jiuhai and Xue, Le and Xu, Zhiyang and Pan, Xichen and Yang, Shusheng and Qin, Can and Yan, An and Zhou, Honglu and Chen, Zeyuan and Huang, Lifu and others},
  journal={arXiv preprint arXiv:2510.15857},
  year={2025}
}

@article{wang2025ovisu1,
  title={Ovis-U1 Technical Report},
  author={Wang, Guo-Hua and Zhao, Shanshan and Zhang, Xinjie and Cao, Liangfu and Zhan, Pengxin and Duan, Lunhao and Lu, Shiyin and Fu, Minghao and Chen, Xiaohao and Zhao, Jianshan and others},
  journal={arXiv preprint arXiv:2506.23044},
  year={2025}
}

@article{deng2025bagel,
  title={Emerging properties in unified multimodal pretraining},
  author={Deng, Chaorui and Zhu, Deyao and Li, Kunchang and Gou, Chenhui and Li, Feng and Wang, Zeyu and Zhong, Shu and Yu, Weihao and Nie, Xiaonan and Song, Ziang and others},
  journal={arXiv preprint arXiv:2505.14683},
  year={2025}
}

@article{wu2025omnigen2,
  title={OmniGen2: Exploration to Advanced Multimodal Generation},
  author={Wu, Chenyuan and Zheng, Pengfei and Yan, Ruiran and Xiao, Shitao and Luo, Xin and Wang, Yueze and Li, Wanli and Jiang, Xiyan and Liu, Yexin and Zhou, Junjie and others},
  journal={arXiv preprint arXiv:2506.18871},
  year={2025}
}

@article{xie2025show,
  title={Show-o2: Improved Native Unified Multimodal Models},
  author={Xie, Jinheng and Yang, Zhenheng and Shou, Mike Zheng},
  journal={arXiv preprint arXiv:2506.15564},
  year={2025}
}

@article{chen2025janus,
  title={Janus-pro: Unified multimodal understanding and generation with data and model scaling},
  author={Chen, Xiaokang and Wu, Zhiyu and Liu, Xingchao and Pan, Zizheng and Liu, Wen and Xie, Zhenda and Yu, Xingkai and Ruan, Chong},
  journal={arXiv preprint arXiv:2501.17811},
  year={2025}
}

@article{kong2024hunyuanvideo,
  title={Hunyuanvideo: A systematic framework for large video generative models},
  author={Kong, Weijie and Tian, Qi and Zhang, Zijian and Min, Rox and Dai, Zuozhuo and Zhou, Jin and Xiong, Jiangfeng and Li, Xin and Wu, Bo and Zhang, Jianwei and others},
  journal={arXiv preprint arXiv:2412.03603},
  year={2024}
}

@article{gao2025seedance,
  title={Seedance 1.0: Exploring the Boundaries of Video Generation Models},
  author={Gao, Yu and Guo, Haoyuan and Hoang, Tuyen and Huang, Weilin and Jiang, Lu and Kong, Fangyuan and Li, Huixia and Li, Jiashi and Li, Liang and Li, Xiaojie and others},
  journal={arXiv preprint arXiv:2506.09113},
  year={2025}
}

@article{liu2024sora,
  title={Sora: A review on background, technology, limitations, and opportunities of large vision models},
  author={Liu, Yixin and Zhang, Kai and Li, Yuan and Yan, Zhiling and Gao, Chujie and Chen, Ruoxi and Yuan, Zhengqing and Huang, Yue and Sun, Hanchi and Gao, Jianfeng and others},
  journal={arXiv preprint arXiv:2402.17177},
  year={2024}
}

@article{chu2023qwenaudio,
  title={Qwen-audio: Advancing universal audio understanding via unified large-scale audio-language models},
  author={Chu, Yunfei and Xu, Jin and Zhou, Xiaohuan and Yang, Qian and Zhang, Shiliang and Yan, Zhijie and Zhou, Chang and Zhou, Jingren},
  journal={arXiv preprint arXiv:2311.07919},
  year={2023}
}

@article{lu2025ovis25technicalreport,
  title={Ovis2.5 Technical Report}, 
  author={Shiyin Lu and Yang Li and Yu Xia and Yuwei Hu and Shanshan Zhao and Yanqing Ma and Zhichao Wei and Yinglun Li and Lunhao Duan and Jianshan Zhao and Yuxuan Han and Haijun Li and Wanying Chen and Junke Tang and Chengkun Hou and Zhixing Du and Tianli Zhou and Wenjie Zhang and Huping Ding and Jiahe Li and Wen Li and Gui Hu and Yiliang Gu and Siran Yang and Jiamang Wang and Hailong Sun and Yibo Wang and Hui Sun and Jinlong Huang and Yuping He and Shengze Shi and Weihong Zhang and Guodong Zheng and Junpeng Jiang and Sensen Gao and Yi-Feng Wu and Sijia Chen and Yuhui Chen and Qing-Guo Chen and Zhao Xu and Weihua Luo and Kaifu Zhang},
  year={2025},
  journal={arXiv:2508.11737}
}

@inproceedings{zhang2019bertscore,
  title={Bertscore: Evaluating text generation with bert},
  author={Zhang, Tianyi and Kishore, Varsha and Wu, Felix and Weinberger, Kilian Q and Artzi, Yoav},
  booktitle={ICLR},
  year={2020}
}

@inproceedings{papineni2002bleu,
  title={Bleu: a method for automatic evaluation of machine translation},
  author={Papineni, Kishore and Roukos, Salim and Ward, Todd and Zhu, Wei-Jing},
  booktitle={ACL},
  year={2002}
}

@incollection{benesty2009pearson,
  title={Pearson correlation coefficient},
  author={Benesty, Jacob and Chen, Jingdong and Huang, Yiteng and Cohen, Israel},
  booktitle={Noise reduction in speech processing},
  year={2009},
}

@misc{DeepMind_Veo3_ModelCard_2025,
  author       = {{Google DeepMind}},
  title        = {Veo 3 Model Card},
  howpublished = {\url{https://storage.googleapis.com/deepmind-media/Model-Cards/Veo-3-Model-Card.pdf}},
  year         = {2025},
  month        = may,
  urldate      = {2025-11-14}
}

@misc{Kuaishou_Kling21_Announcement_2025,
  author       = {{Kuaishou Technology}},
  title        = {Kling AI},
  howpublished = {\url{https://klingai.kuaishou.com/}},
  year         = {2025}
}

@article{li2025omnivideobench,
  title={Omnivideobench: Towards audio-visual understanding evaluation for omni mllms},
  author={Li, Caorui and Chen, Yu and Ji, Yiyan and Xu, Jin and Cui, Zhenyu and Li, Shihao and Zhang, Yuanxing and Tang, Jiafu and Song, Zhenghao and Zhang, Dingling and others},
  journal={arXiv preprint arXiv:2510.10689},
  year={2025}
}

@article{lin2025sail,
  title={SAIL-Embedding Technical Report: Omni-modal Embedding Foundation Model},
  author={Lin, Lin and Long, Jiefeng and Wan, Zhihe and Wang, Yuchi and Yang, Dingkang and Yang, Shuang and Yao, Yueyang and Chen, Xu and Guo, Zirui and Li, Shengqiang and others},
  journal={arXiv preprint arXiv:2510.12709},
  year={2025}
}

@article{liu2025reinforcement,
  title={Reinforcement learning meets large language models: A survey of advancements and applications across the llm lifecycle},
  author={Liu, Keliang and Yang, Dingkang and Qian, Ziyun and Yin, Weijie and Wang, Yuchi and Li, Hongsheng and Liu, Jun and Zhai, Peng and Liu, Yang and Zhang, Lihua},
  journal={arXiv preprint arXiv:2509.16679},
  year={2025}
}

@article{yang2025medaide,
  title={Medaide: Information fusion and anatomy of medical intents via llm-based agent collaboration},
  author={Yang, Dingkang and Wei, Jinjie and Li, Mingcheng and Liu, Jiyao and Liu, Lihao and Hu, Ming and He, Junjun and Ju, Yakun and Zhou, Wei and Liu, Yang and others},
  journal={Information Fusion},
  pages={103743},
  year={2025},
  publisher={Elsevier}
}

@inproceedings{yang2024pediatricsgpt,
  title={Pediatricsgpt: Large language models as chinese medical assistants for pediatric applications},
  author={Yang, Dingkang and Wei, Jinjie and Xiao, Dongling and Wang, Shunli and Wu, Tong and Li, Gang and Li, Mingcheng and Wang, Shuaibing and Chen, Jiawei and Jiang, Yue and others},
  booktitle={NeurIPS},
  year={2024}
}

@article{yang2025improvingmsa,
  title={Improving Multimodal Sentiment Analysis via Modality Optimization and Dynamic Primary Modality Selection},
  author={Yang, Dingkang and Li, Mingcheng and Wu, Xuecheng and Chen, Zhaoyu and Jiang, Kaixun and Liu, Keliang and Zhai, Peng and Zhang, Lihua},
  journal={arXiv preprint arXiv:2511.06328},
  year={2025}
}

@inproceedings{yang2022disentangled,
author = {Yang, Dingkang and Huang, Shuai and Kuang, Haopeng and Du, Yangtao and Zhang, Lihua},
title = {Disentangled Representation Learning for Multimodal Emotion Recognition},
booktitle={ACM MM},
year = {2022}
}

@article{yang2024towards,
  title={Towards context-aware emotion recognition debiasing from a causal demystification perspective via de-confounded training},
  author={Yang, Dingkang and Yang, Kun and Kuang, Haopeng and Chen, Zhaoyu and Wang, Yuzheng and Zhang, Lihua},
  journal={IEEE Transactions on Pattern Analysis and Machine Intelligence},
  year={2024},
  publisher={IEEE}
}

@article{yang2024asynchronous,
  title={Asynchronous Multimodal Video Sequence Fusion via Learning Modality-Exclusive and-Agnostic Representations},
  author={Yang, Dingkang and Li, Mingcheng and Qu, Linhao and Yang, Kun and Zhai, Peng and Wang, Song and Zhang, Lihua},
  journal={IEEE Transactions on Circuits and Systems for Video Technology},
  year={2024},
  publisher={IEEE}
}

\clearpage


\end{document}